\documentclass{article} 
\usepackage{iclr2026_conference,times}
\usepackage{enumitem}
\usepackage{wrapfig}
\usepackage{float}
\usepackage{marvosym}
\iclrfinalcopy
\usepackage{xcolor}
\usepackage{colortbl}
\definecolor{PTDRowBackground}{HTML}{EAF2FA}

\usepackage{amsmath,amsfonts,bm}
\usepackage{graphicx}
\usepackage{booktabs}
\usepackage{multirow}
\usepackage{graphicx}

\def\eqref#1{equation~\ref{#1}}

\def\1{\bm{1}}

\DeclareMathAlphabet{\mathsfit}{\encodingdefault}{\sfdefault}{m}{sl}
\SetMathAlphabet{\mathsfit}{bold}{\encodingdefault}{\sfdefault}{bx}{n}

\usepackage{hyperref}
\usepackage{url}
\usepackage{cleveref}
\crefname{equation}{Eq.}{}

\title{Teaching the Way, Not the Answer: Privileged Tutoring Distillation for Multimodal Policy Optimization}

\author{%
\textbf{Shizhe Xiang}\textsuperscript{1*\dag},
\textbf{Ke An}\textsuperscript{2*\dag},
\textbf{Wenlong Yu}\textsuperscript{3},
\textbf{Yue Liu}\textsuperscript{4\dag},\\
\textbf{Tingting Wu}\textsuperscript{1},
\textbf{Jian Luan}\textsuperscript{5},
\textbf{Pei Fu}\textsuperscript{5\Letter},
\textbf{Qilong Wang}\textsuperscript{1\Letter}\\
\textsuperscript{1}Tianjin University \quad
\textsuperscript{2}Beijing Institute of Technology \\
\textsuperscript{3}Singapore Management University \quad
\textsuperscript{4}University of Chinese Academy of Sciences \\
\textsuperscript{5}Xiaomi Inc.
}

\begin{document}

\maketitle

\begingroup
\renewcommand{\thefootnote}{\fnsymbol{footnote}}
\footnotetext[1]{Equal contribution.}
\footnotetext[2]{Work done during the internship at Xiaomi.}
\renewcommand{\thefootnote}{\Letter}
\footnotetext[3]{Qilong Wang: primary corresponding author; Pei Fu: co-corresponding author.}
\endgroup

%

\newcommand{\fix}{\marginpar{FIX}}
\newcommand{\new}{\marginpar{NEW}}


\begin{abstract}
Recent post-training methods, particularly Reinforcement Learning with Verifiable Rewards (RLVR), have significantly enhanced the reasoning ability of Large Vision-Language Models (LVLMs). However, the sparse nature of verifiable rewards provides little token-level supervision for failed rollouts, often leading to inefficient exploration in complex multimodal reasoning tasks. Although policy distillation can offer dense guidance, external teacher based methods introduce substantial computational overhead, while answer conditioned tuning methods potentially expose answer-level information and induce shortcut-like generation behavior. To overcome above limitations, we propose a novel Privileged Tutoring Distillation Policy Optimization (PTD-PO) method for RLVR, whose core idea is to supplement supervision with guidance on the reasoning path rather than the answer itself. Specifically, our PTD-PO first constructs Way-revealing Structured Privileged Hints by considering both spatial attention guidance and intermediate textual reasoning steps, which are further employed to guide the student model toward corrective reasoning paths through in-context-learning. particularly, the student model is still optimized under the original question-only context, while only failed rollouts are aligned with the hint-augmented reference model at the token-distribution level. To further stabilize distillation under the distribution shift between guided and unguided contexts, we introduce a Top-K Jensen-Shannon divergence with tail compensation, focusing on aligning informative tokens while reducing memory overhead. Experiments on LVLMs ranging from 2B to 8B parameters show that our PTD-PO can provide effective dense guidance without relying on answer-revealing supervision and consistently outperforms state-of-the-art counterparts by a clear margin.
Code is available at \url{https://github.com/XszNeverSleep/PTD-PO}.
\end{abstract}

\section{Introduction}

Recent advances in Large Vision-Language Models (LVLMs) have enabled strong performance on complex multimodal reasoning tasks, where models need to ground visual evidence, perform multi-step inference, and produce verifiable answers \citep{liu2023visual, yue2024mmmu, lu2023mathvista, bai2025qwen3, dai2023instructblip, li2023blip}. Reinforcement Learning with Verifiable Rewards (RLVR) has become a promising post-training paradigm for further eliciting these above abilities, as it replaces costly human preference annotation with outcome-level feedback \citep{ouyang2022training, shao2024deepseekmath, guo2025deepseek, team2025kimi, askell2021general}. However, verifiable rewards are usually assigned only to the final answer   \citep{lightman2023let, chen2024we}, where failed rollouts provide little information on incorrect intermediate visual grounding or reasoning steps. Therefore, it makes credit assignment difficult and often cause exploration failure in large reasoning spaces \citep{prakash2025can, tran2025exploiting, ren2026recycling}.

A natural way to alleviate reward sparsity is replacement or supplement of outcome-level RLVR with dense token-level guidance \citep{agarwal2024policy, gu2024minillm}. On-Policy Distillation (OPD) provides such guidance by aligning student rollouts with a teacher policy during training. However, it typically requires online inference from an external teacher over student-generated trajectories, incurring substantial computational overhead, lowering training efficiency, and potentially introducing tokenizer or vocabulary mismatch \citep{fu2026revisiting, minixhofer2025cross}. A more efficient alternative is contextual self-distillation, which constructs the teacher from the student model itself under additional conditioning \citep{zhao2026self, hubotter2026reinforcement}. Generally, existing methods directly condition the self-teacher on ground-truth answers or complete solution traces to obtain strong supervision. Although advanced efforts are made, we argue that such answer-revealing contexts leads to ineffective teaching behavior, where potentially may bias the teacher toward answer-level shortcuts and sharpened token distributions,  rather than guiding the student model to discover the reasoning path and so reducing diversity of reasoning trajectories. The similar phenomenon on degraded exploration was also observed in prior work \citep{zhang2026heal, yang2026self}. Therefore, a question is naturally rasied up: Can we provide the effective yet efficient dense supervisions on multimodal policy optimization without answer-revealing contexts  and costly online external teachers?

\begin{figure}[t]
\vspace{-12pt}
  \centering
  \includegraphics[width=\linewidth, height=0.9\textheight, keepaspectratio]{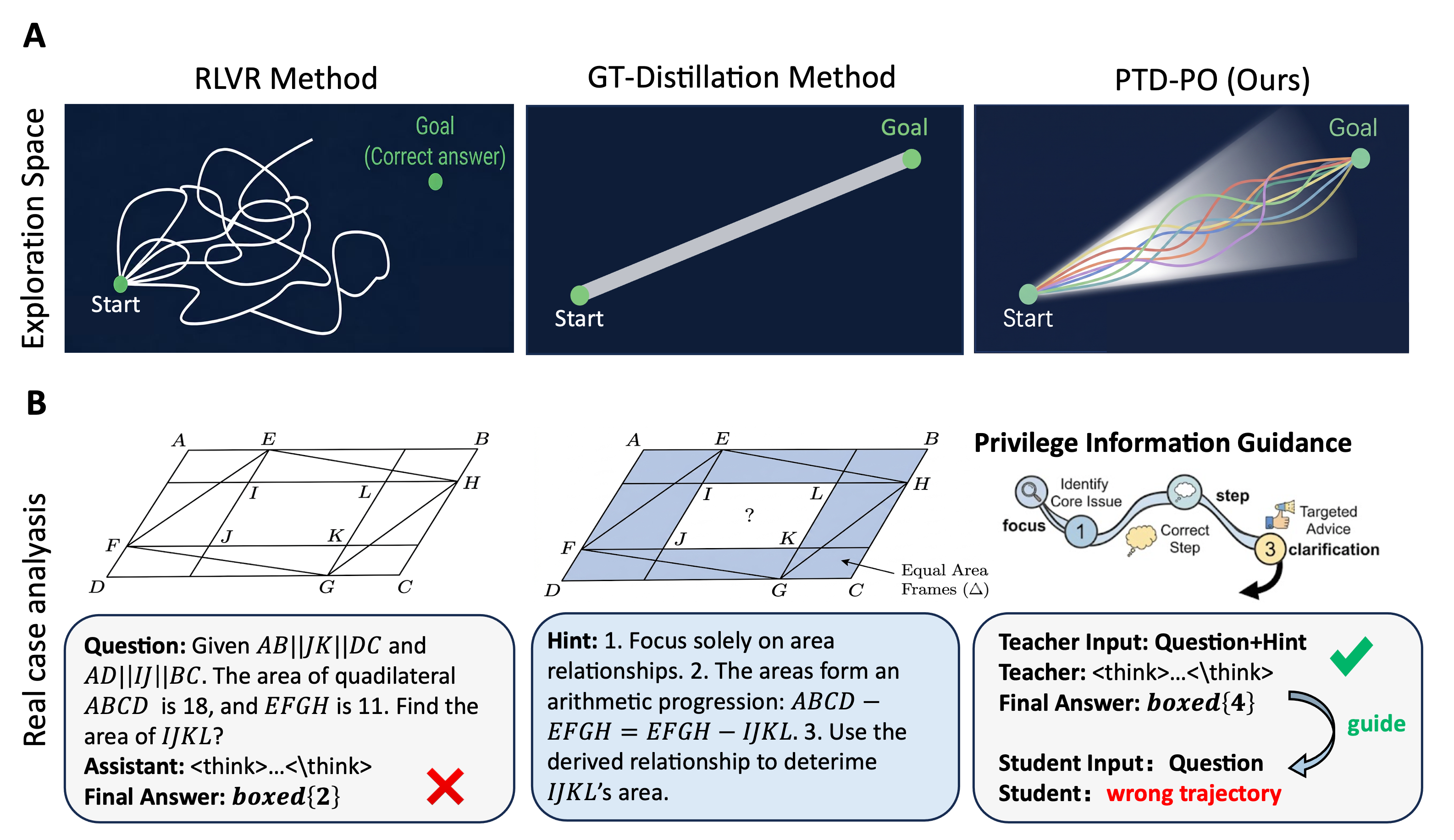}
  \vspace{-12pt}
\caption{
(a) PTD-PO complements sparse-reward RLVR with way-revealing structured privileged hints, avoiding the overly deterministic behavior of answer-revealing distillation.
(b) Given a failed geometry example, the actor gives a wrong answer from the question-only context.
The way-revealing hint guides the reference model to focus on area relationships and equal-area frames, and to infer the relation among $ABCD$, $EFGH$, and $IJKL$ without exposing the final area. The hint-augmented teacher then provides dense corrective supervision for the failed trajectory.
}
  \label{fig:intro_img}
  \vspace{-12pt}
\end{figure}


To address this question, we propose Privileged Tutoring Distillation Policy Optimization (PTD-PO), inspired by an ancient Taoist proverb\footnote{Give a man a fish and you feed him for a day. Teach a man to fish and you feed him for a lifetime.}. As illustrated in Figure~\ref{fig:intro_img}, PTD-PO uses way-revealing hints to guide a reference model in correcting failed student trajectories, while the student remains conditioned only on the original question. Our PTD-PO is developed with two key components, i.e., Way-revealing Structured Privileged Hint (WSPH) and failure-routed Privileged Tutoring Distillation (PTD), under the learning-with-privileged-information paradigm \citep{vapnik2009new,sharmanska2013learning}. Specifically, WSPH construction combines where-to-look visual grounding with how-to-reason guidance, followed by refinement to remove answer-revealing or unreliable hints. Based on WSPH, PTD routes tutoring signals to failed rollouts and aligns tutor and student token distributions under asymmetric contexts, enabling targeted supervision for reasoning recovery while preserving the student's original problem context. Furthermore, we introduce a Top-K Jensen-Shannon Divergence with tail compensation in PTD to efficiently align informative token distributions while accounting for the remaining probability mass under long-context reasoning. As such, our PTD-PO can transform sparse outcome feedback into targeted reasoning-path supervision through, teaching the student how to recover from failures rather than imitate answers, thereby improving training stability and generalization. We evaluate PTD-PO on LVLMs ranging from 2B to 8B parameters across complex multimodal reasoning benchmarks\citep{wang2025perception}, clearly demonstrating its ability to provide dense reasoning guidance while avoiding answer-revealing
supervision during RLVR post-training.

In summary, our contributions are as follows:
\begin{itemize}[leftmargin=*, nosep]
\item We disclose the issues of ineffective teaching behavior  in sparse outcome rewards and answer-revealing self-distillation for multimodal policy optimization. They encourage us to propose an effective privileged tutoring distillation (PTD-PO) with core idea of multimodal policy optimization with way-revealing tutoring rather than answer-revealing self-distillation.

\item To this end, our PTD-PO presents Way-revealing Structured Privileged Hint (WSPH) and failure-routed Privileged Tutoring Distillation (PTD) with Top-K Jensen-Shannon Divergence with tail compensation, which routes dense supervision to failed trajectories by conditioning the reference model on way-revealing hints of both visual evidence and text reasoning, efficiently transforming sparse outcome feedback into dense yet informative reasoning-path supervision. 

\item Extensive experiments show that PTD-PO consistently improves LVLM performance across model scales and a diverse range of multimodal reasoning benchmarks. Furthermore, our PTD-PO can be seamlessly integrated with different RLVR optimizers, highlighting its strong generalization ability.
\end{itemize}

\section{Methodology}
\subsection{Preliminary}



\paragraph{Policy Optimization}
Policy optimization is a fundamental paradigm for multimodal model post-training, where the policy is optimized using feedback from sampled responses. Group Relative Policy Optimization (GRPO) as one of most popular methods \citep{shao2024deepseekmath} leverages group-relative rewards without requiring an explicit value function. Given a multimodal question \(x\), GRPO samples a group of \(G\) responses \(\{y_i\}_{i=1}^{G}\) with \(y_i=(y_{i,1},\ldots,y_{i,T_i})\), where \(T_i\) denotes the response length. Each response receives a verifiable reward \(r_i\), from which the group-relative advantage is computed as $A_i=
    \frac{r_i-\mathrm{mean}(\{r_j\}_{j=1}^{G})}
    {\mathrm{std}(\{r_j\}_{j=1}^{G})+\epsilon}$, where \(\epsilon\) is a small scalar for numerical stability. Let \(\pi_{\theta_{\mathrm{old}}}\) denote the rollout policy, for each token \(y_{i,t}\) with prefix \(y_{i,<t}\), GRPO defines $\rho_{i,t}(\theta)=\pi_\theta(y_{i,t}\mid x,y_{i,<t})/\pi_{\theta_{\mathrm{old}}}(y_{i,t}\mid x,y_{i,<t})$ and optimizes the clipped objective:
\begin{equation}
\label{eq:GRPO}
    J_{\mathrm{GRPO}}(\theta)
    =
    \mathbb{E}_{x,\{y_i\}_{i=1}^{G}}
    \left[
    \frac{1}{G}\sum_{i=1}^{G}\frac{1}{T_i}
    \sum_{t=1}^{T_i}
    \left(
    \min\!\left(
    \rho_{i,t}(\theta)A_i,
    \mathrm{clip}(\rho_{i,t}(\theta),1-\epsilon_c,1+\epsilon_c)A_i
    \right)
    \right)
    \right],
\end{equation}
where \(\epsilon_c\) is a clipping coefficient. While GRPO avoids an explicit value function, its learning signal remains primarily outcome-level. When all sampled responses fail, it provides limited guidance for correcting intermediate reasoning steps. This motivates on-policy distillation to complement answer-level feedback with dense token-level supervision.

\paragraph{On-Policy Distillation and Self-Distillation.}
On-policy distillation \citep{ye2026policy} provides dense token-level supervision by aligning the student policy $\pi_\theta$ with a teacher policy $\pi_T$. Given an input $x$ and a rollout $y_i$ sampled from the student, where $s_{i,t}=(x,y_{i,<t})$ denotes the tokens generated before step $t$. The standard online policy distillation objective is define as:
\begin{equation}
\mathcal{J}_{\mathrm{OPD}}(\theta)
=
\mathbb{E}_{x,\{y_i\}}
\left[
\frac{1}{G}\sum_{i=1}^{G}\frac{1}{T_i}\sum_{t=1}^{T_i}
D_{\mathrm{KL}}
\left(
\pi_T(\cdot\mid s_{i,t})
\|
\pi_\theta(\cdot\mid s_{i,t})
\right)
\right].
\label{eq:opd_objective}
\end{equation}
Compared with GRPO (\cref{eq:GRPO}) that only provides scalar outcome-level rewards, OPD in \cref{eq:opd_objective} specifies a token-level target distribution at each reasoning step and thus offers more informative optimization signals. However, when $\pi_T$ is instantiated as a larger external model, on-policy distillation requires additional teacher inference and  potentially suffers from tokenizer or vocabulary mismatch.

To avoid the cost and dependency of external teachers, self-distillation \citep{agarwal2024policy} instantiates the teacher policy \(\pi_T\) in the above policy distillation objective using the actor model itself, where SDPO \citep{hubotter2026reinforcement} uses the exponential moving average of the actor parameters, \(\mathrm{EMA}(\pi_\theta)\), for \(\pi_T\), while OPSD \citep{zhao2026self} uses the frozen reference model \( (\pi_{\mathrm{ref}})\) for \(\pi_T\). Although this removes the need for an external teacher, the corrective supervision provided by self-distillation depends on the additional context supplied to the teacher. We analyze this dependence below.

\subsection{Analysis on limitations of answer-revealing policy optimization}
\begin{wrapfigure}{r}{0.56\textwidth}
    \vspace{-10pt}
    \centering
    \includegraphics[width=\linewidth]{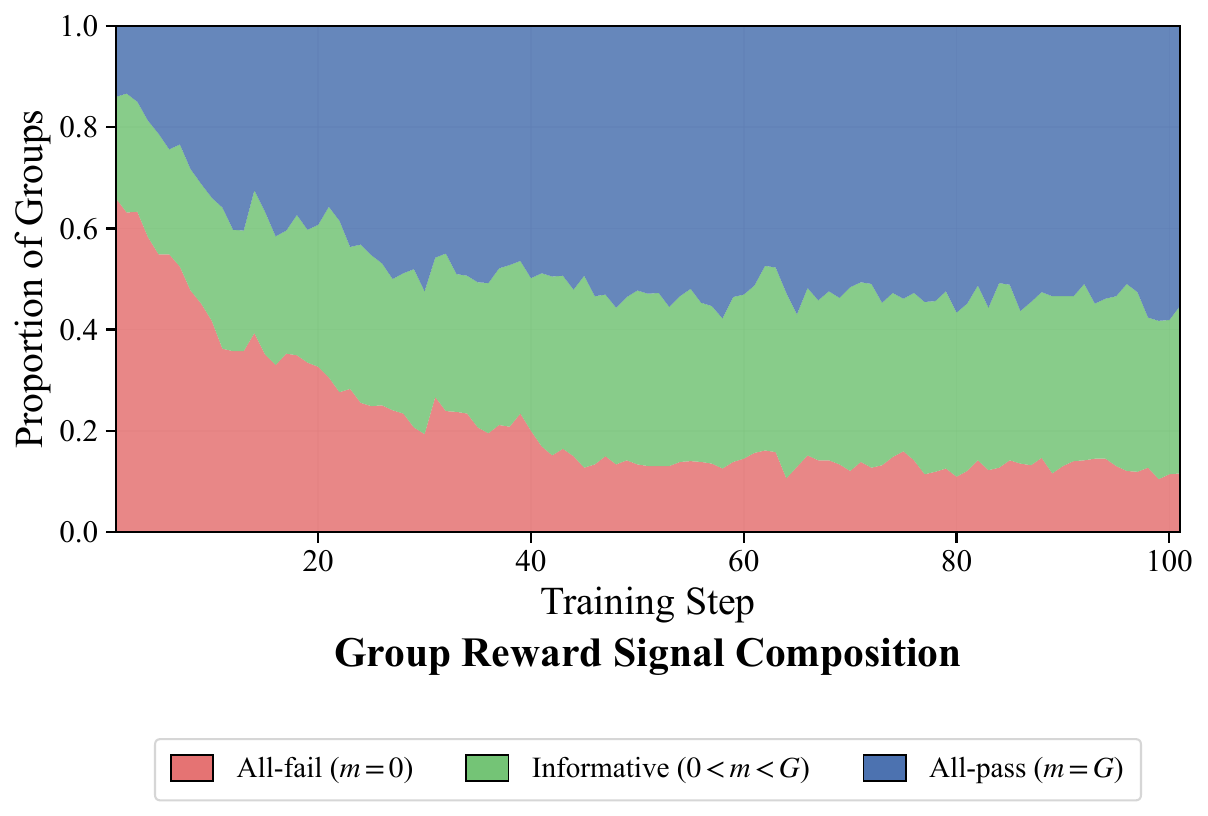}
    \vspace{-8pt}
    \caption{Group-level reward composition during GRPO training. Informative groups $(0<m<G)$ remain limited, while all-fail and all-pass groups yield homogeneous rewards.}
    \label{fig:group_reward_stats}
    \vspace{-10pt}
\end{wrapfigure}

\paragraph{Analysis 1: Limited Informative Reward Groups in Answer-Only Policy Optimization.}
Answer-only policy optimization relies on outcome-level feedback derived from the correctness of sampled responses. To examine the informativeness of such supervision, we analyze the group-level reward composition in GRPO as a representative example. For each question $x$, we sample $G$ responses and categorize each group by the number of successful responses, $m(x)=\sum_{i=1}^{G}r_i$, into all-fail $(m=0)$, informative $(0<m<G)$, and all-pass $(m=G)$ groups. As shown in Figure~\ref{fig:group_reward_stats}, all-fail groups dominate the early training stage, while informative groups remain limited to roughly one third of all sampled groups and all-pass groups gradually become dominant. This indicates that answer-only policy optimization frequently provides homogeneous or low-information reward signals, limiting its effectiveness for correcting failed trajectories. This observation shares similar philosophy with recent studies \citep{he2026advantage,yu2026dapo}, showing that homogeneous rewards within a group lead to advantage collapse or near-zero learning signals in GRPO.

\begin{figure}[H]
    \centering
    \makebox[\linewidth][c]{%
    \includegraphics[
        width=1.03\linewidth,
        trim=0 0 0 6bp,
        clip
    ]{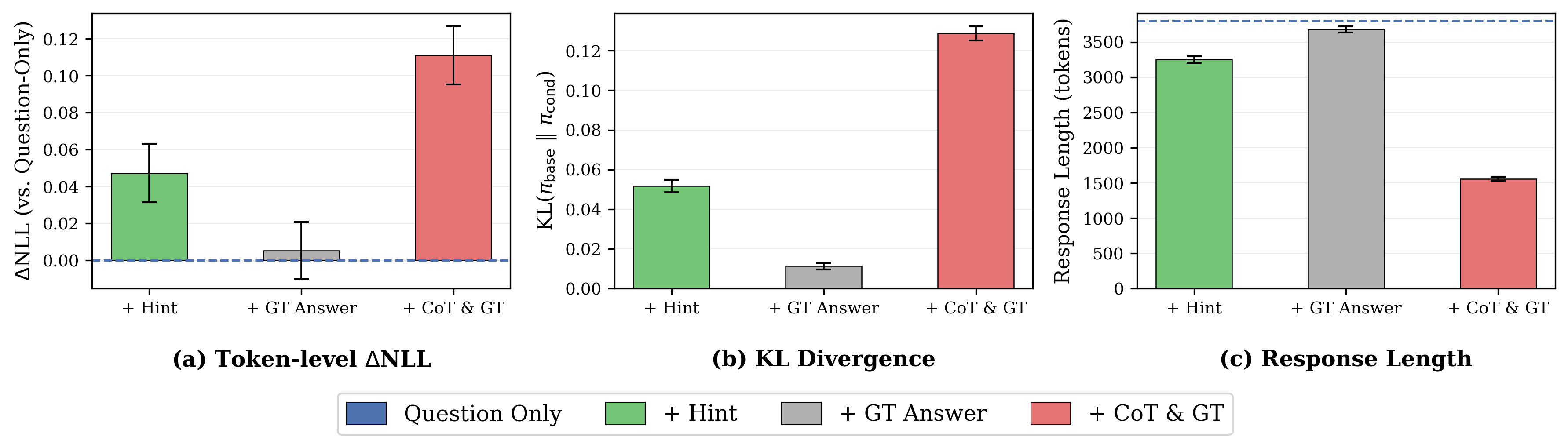}
    }
    \caption{
    Context-conditioned self-teacher statistics.
    (a) Token-level $\Delta$NLL, (b) full-distribution KL divergence, and (c) response length show that complete solution traces induce the largest distribution shift and shortcut-like generation behavior, while our Way-revealing Structured Privileged Hint context provide a moderate guidance signal.
    }
    \label{fig:teacher_context_shift}
\end{figure}


\paragraph{Analysis 2: Answer-Revealing Contexts Induce Distribution Shift and Shortcut-Like Teacher Behavior.}
We further examine how answer-revealing information changes the token distribution of a frozen self-teacher. All conditions use the same self-teacher and differ only in context. As shown in Figure~\ref{fig:teacher_context_shift}(a) and (b), providing only the final GT answer has a marginal effect on the self-teacher distribution, where both token-level $\Delta$NLL and full-distribution KL remain close to the question-only baseline. In contrast, conditioning the self-teacher on both the complete CoT solution and GT answer induces a much larger distribution shift, with substantially higher $\Delta$NLL on student-generated tokens and the largest KL divergence. Moreover, as shown in Figure~\ref{fig:teacher_context_shift}(c), the CoT-and-GT condition sharply reduces response length, suggesting that the self-teacher increasingly relies on the provided solution rather than autonomous reasoning. These results highlight a concern with answer-revealing supervision: additional answer information can alter the self-teacher's reasoning behavior and induce shortcut-like generation. Above observations encourage us to provide an alternative effective guidance for policy optimization without exposing complete answer-level information. Therefore, we propose a Privileged Tutoring Distillation Policy Optimization (PTD-PO) method to perfrom policy optimization with way-revealing tutoring rather than answer-revealing self-distillation, whose details are described as follows.

\subsection{PTD-PO: Privileged Tutoring Distillation Policy Optimization}

\begin{figure*}[t]
    \centering
    \includegraphics[width=\textwidth]{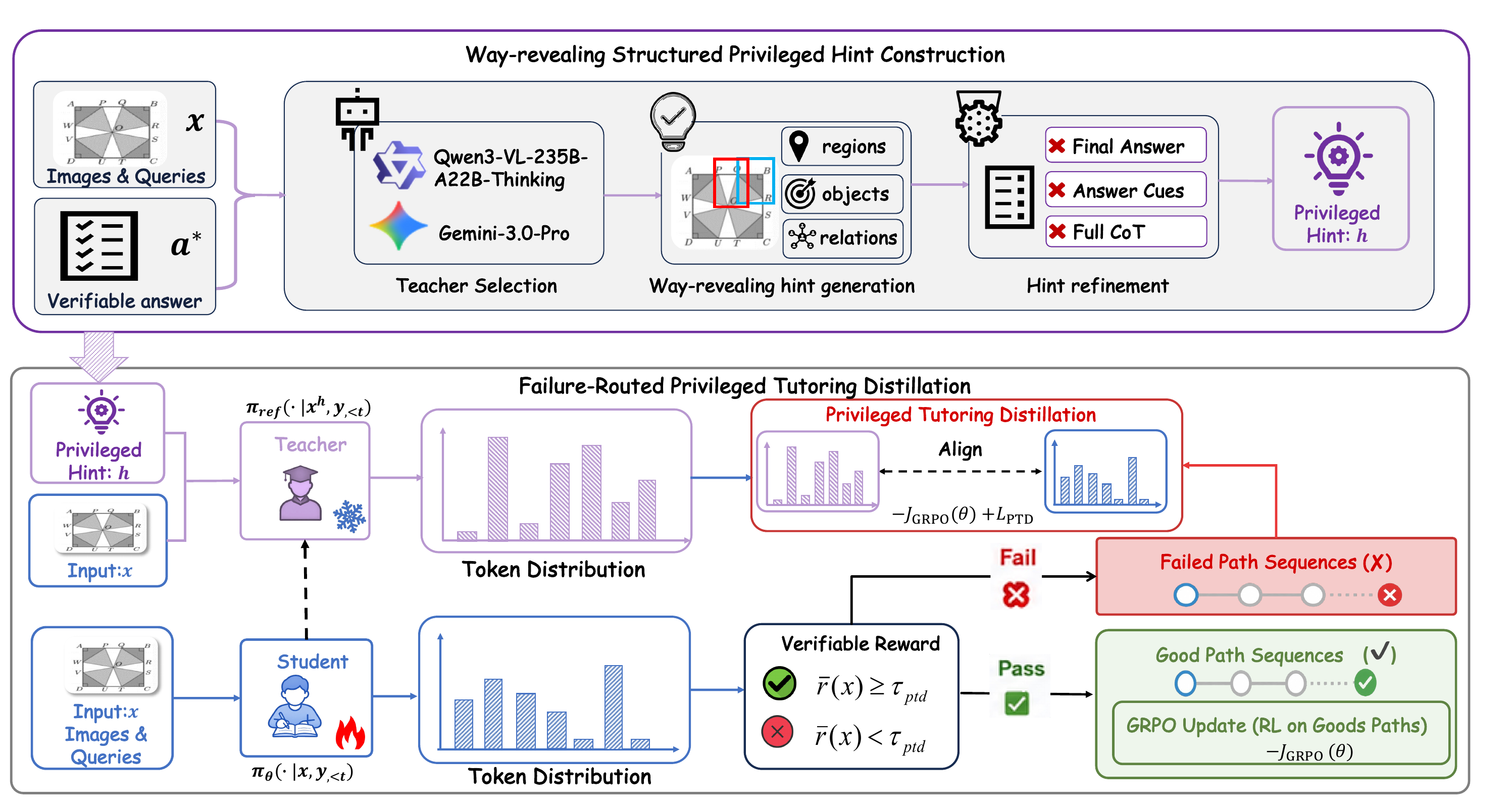}
    \caption{
    Overview of PTD-PO, which combines RLVR with privileged-tutoring distillation from the Way-revealing Privileged Hint conditioned reference model on failed trajectories.
    }
    \label{fig:pid_grpo_arch}
\end{figure*}


To address the limited supervision for failed RLVR trajectorie, PTD-PO is proposed by turning them into targets for corrective tutoring while preserving exploration. As shown in Figure~\ref{fig:pid_grpo_arch}, our PTD-PO consists of two key components: Way-revealing Structured Privileged Hint (WSPH) construction and failure-routed Privileged Tutoring Distillation (PTD). Specifically, the actor samples responses under the original prompt and receives verifiable rewards to identify failed trajectories. Building on this, WSPH provides structured spatial and reasoning hints for these trajectories, while PTD routes the tutoring signals by aligning tutor--student token distributions under asymmetric contexts. The resulting signal complements the reward-driven objective to update the actor, turning failed rollouts into dense supervision for learning reasoning strategies rather than answer-conditioned shortcuts. The following sections detail how WSPH constructs non-spoiling guidance and how PTD effectively distills it into the actor.

\paragraph{Way-revealing Structured Privileged Hint Construction.}
\label{par:wsph}
For each training example, we construct a structured privileged hint from the multimodal question \(x\) and verifiable answer \(a^\star\), where answer is only used to construct a way-revealing hint and is never exposed to the student policy.
The hint provides only partial, process-level guidance, thereby inducing a milder privileged-context shift from the original question. To preserve this property, the hint is constrained to indicate a valid reasoning direction without revealing answer-identifying information.


Specifically, we use Qwen3-VL-235B-A22B-Thinking as the primary external teacher and Gemini-3.0-Pro as a fallback model to generate the hints. The construction consists of two stages: way-revealing hint generation and hint refinement. Given \((x,a^\star)\), a rule-guided system prompt instructs the teacher to identify task-relevant visual evidence, including key regions, objects, and spatial relations, and formulate a concise and informative reasoning direction based on these cues. We then perform lightweight hint refinement to filter invalid candidates, including those with answer leakage, missing or irrelevant visual evidence, or overly complete reasoning traces. The resulting validated hints serve as the final privileged annotations for subsequent policy training. Figure~\ref{fig:hint_ablation} (a) illustrates a representative construction example, while the complete prompt, refinement rules, and additional cases are provided in Appendices~\ref{app:hints_construction} and~\ref{app:case_study}.

\paragraph{Failure-Routed Privileged Tutoring Distillation.}
Building on WSPH, we incorporate the
way-revealing hints into on-policy distillation to provide effective supervision on corrective lines of reasoning for failed trajectories. Given a multimodal question $x$, the rollout policy samples a group of $G$ responses
\(\{y_i\}_{i=1}^{G}\) under the original question-only context, i.e.,
\begin{equation}
y_i \sim \pi_{\theta_{\mathrm{old}}}(\cdot \mid x),
\qquad
r_i = R(x, y_i) \in \{0,1\},
\end{equation}
where $R(\cdot)$ is the verifiable reward function. We characterize the difficulty and
training informativeness of the sampled group by its empirical accuracy, i.e.,
\begin{equation}
\bar r(x) = \frac{1}{G}\sum_{i=1}^{G} r_i ,
\end{equation}
where small $\bar r(x)$ indicates most trajectories fail to reach the correct answer,
and the answer-level reward provides limited information about how their intermediate
visual grounding or reasoning steps should be corrected. Therefore, instead of applying
privileged tutoring distillation uniformly to all rollouts, we route PTD supervision to the
incorrect responses in the sampled group. Specifically, given an activation threshold
$\tau_{\mathrm{ptd}}$, the set of token positions selected for PTD is defined as:
\begin{equation}
\mathcal{F}
=
\left\{
(i,t) \;\middle|\;
\bar{r}(x) < \tau_{\mathrm{ptd}},\;
r_i = 0,\;
m_{i,t}=1
\right\},
\end{equation}
where $m_{i,t}$ is the response mask. Particularly,  we use
$\tau_{\mathrm{ptd}}=1.0$ as the default setting, indicating that PTD is applied to all failed trajectories.

For each question $x$, we construct a way-revealing privileged hint $h$ and form the
hint-augmented teacher context $x^h=[x;h]$. In the OPD objective of \cref{eq:opd_objective}, we replace the teacher distribution
$\pi_T(\cdot\mid s_{i,t})$ with the hint-conditioned frozen reference distribution
$q^h_{i,t}=\pi_{\mathrm{ref}}(\cdot\mid x^h,y_{i,<t})$, while retaining the
question-only student distribution
$p_{i,t}=\pi_\theta(\cdot\mid x,y_{i,<t})$.
This asymmetric construction decouples the information used to produce dense supervision
from the information available to the student policy. As such, the teacher policy can use privileged
spatial and textual hints to infer a corrective reasoning direction, whereas the student
must learn to reproduce the improved token distribution without directly observing the
hint. To this end, our privileged tutoring distillation loss is defined over the selected failed positions, i.e.,
\begin{equation}
\mathcal{L}_{\mathrm{PTD}}
=
\frac{
\sum_{(i,t)\in \mathcal{F}}
D_{\mathrm{PTD}}\left(q^h_{i,t}, p_{i,t}\right)
}{
|\mathcal{F}|+\epsilon
} ,
\label{eq:PTD}
\end{equation}
where $D_{\mathrm{PTD}}$ measures the discrepancy between the hint-conditioned teacher distribution and the question-only student distribution. Since the two distributions are conditioned on different contexts, a directional KL divergence potentially introduce a one-sided matching bias, particularly when the teacher distribution is sharp. We therefore base $D_{\mathrm{PTD}}$ on the symmetric and bounded Jensen--Shannon divergence (JSD). For distributions $q$ and $p$, the full JSD is defined as $D_{\mathrm{JSD}}(q,p)
= \frac{1}{2}D_{\mathrm{KL}}(q\Vert m)
+ \frac{1}{2}D_{\mathrm{KL}}(p\Vert m)$, where  $m=\frac{1}{2}(q+p)$.

Finally, we perform policy optimization by combining the GRPO with our PTD regularizer:
\begin{equation}
\mathcal{L}_{\mathrm{PTD\text{-}PO}}
=-J_{\mathrm{GRPO}}(\theta)
+
\lambda_{\mathrm{PTD}}
\mathcal{L}_{\mathrm{PTD}},
\end{equation}
where $\lambda_{\mathrm{PTD}}$ controls the effect of privileged distillation.

\paragraph{Top-$K$ JSD with Tail Compensation.}
Obviously, retaining full-vocabulary distributions at every response position is
memory-intensive. 
To reduce this overhead while preserving the dominant probability mass, we approximate the full JSD with a compact Top-$K$ support and explicitly compensate for the remaining tail mass. At
each selected position $(i,t)$, we take the union of the teacher and student Top-$K$
tokens, retaining high-probability tokens favored by either distribution:
\begin{equation}
\mathcal{S}_{i,t}
=
\operatorname{Top\text{-}K}(p_{i,t})
\cup
\operatorname{Top\text{-}K}(q^h_{i,t}).
\end{equation}
Then, the probability mass outside this support is aggregated into a single tail bucket:
\begin{equation}
p^{\mathrm{tail}}_{i,t}
=
1-\sum_{v\in \mathcal{S}_{i,t}} p_{i,t}(v),
\qquad
q^{h,\mathrm{tail}}_{i,t}
=
1-\sum_{v\in \mathcal{S}_{i,t}} q^h_{i,t}(v).
\end{equation}
\begin{table*}[t]
\centering
\caption{
Comparison of different training methods with various LVLMs on general multimodal reasoning and vision-dependent multimodal reasoning tasks.
}
\label{tab:main_results}

\Large
\renewcommand{\arraystretch}{1.2}
\setlength{\tabcolsep}{1.6mm}

\resizebox{\textwidth}{!}{%
\begin{tabular}{l|cccccc|ccccc|c}
\toprule
\multirow{2}{*}{\textbf{Method / Setting}} 
& \multicolumn{6}{c|}{\textbf{General Multimodal Reasoning}} 
& \multicolumn{5}{c|}{\textbf{Vision-Dependent Multimodal Reasoning}} 
& \multicolumn{1}{c}{\textbf{Overall}} \\
\cmidrule(lr){2-7} \cmidrule(lr){8-12} \cmidrule(lr){13-13}
& \textbf{MMK12} 
& \textbf{Geo3K} 
& \textbf{MathVerse} 
& \textbf{MathVista} 
& \textbf{We-Math} 
& \textbf{AVG} 
& \textbf{MMMU-Pro} 
& \textbf{Counting} 
& \textbf{MathVerse$_V$} 
& \textbf{LogicVista} 
& \textbf{AVG} 
& \textbf{AVG} \\
\midrule

\multicolumn{13}{c}{\textbf{Qwen3-VL-2B-Thinking}} \\
\midrule
SFT 
& 38.59 & 41.95 & 55.50 & 49.05 & 59.55 & 48.93
& 20.56 & 53.44 & 51.25 & 36.83 & 40.52
& 45.19 \\

OPSD
& 44.99 & 38.48 & 52.69 & 48.52 & 56.90 & 48.32
& 17.38 & 61.19 & 46.41 & 30.20 & 38.80
& 44.08 \\

GRPO 
& 51.30 & 45.36 & 53.49 & 56.14 & 58.36 & 52.93
& 23.31 & 84.50 & 49.91 & 33.31 & 47.76
& 50.63 \\

HDPO
& 50.22 & 49.36 & 66.60 & 58.86 & 69.81 & 58.97
& 29.10 & 83.38 & 63.21 & 48.55 & 56.06
& 57.68 \\

PAPO 
& 49.78 & 44.64 & 60.00 & 56.60 & 64.27 & 55.06 
& 24.88 & 88.38 & 54.78 & 43.70 & 
52.94 & 54.11 \\

\rowcolor{PTDRowBackground}
\textbf{PTD-PO(Ours)}
& \textbf{56.86} & \textbf{50.52} & \textbf{70.53} & \textbf{60.38} & \textbf{72.64} & \textbf{62.19}
& \textbf{32.26} & \textbf{88.88} & \textbf{66.62} & \textbf{52.21} & \textbf{59.99}
& \textbf{61.21} \\

\midrule

\multicolumn{13}{c}{\textbf{Qwen3-VL-4B-Thinking}} \\
\midrule
SFT 
& 59.59 & 57.38 & 70.85 & 57.07 & 71.24 & 63.23
& 29.91 & 72.38 & 66.39 & 46.50 & 53.80
& 59.03 \\

OPSD  
& 60.97 & 58.13 & 67.98 & 55.23 & 68.06 & 62.07
& 23.20 & 72.19 & 67.87 & 41.50 & 51.19
& 57.24 \\

GRPO 
& 64.53 & 62.62 & 75.25 & 68.38 & 78.30 & 69.82
& 40.19 & 91.69 & 73.03 & 58.56 & 65.87
& 68.06 \\

HDPO
& 66.26 & 62.75 & 76.25 & 67.83 & 76.93 & 70.00
& 39.93 & 91.75 & 73.84 & 59.06 & 66.15
& 68.29 \\

PAPO 
& 65.11 & 62.69 & 77.73 & \textbf{70.95} & 80.32 & 71.36
& 40.03 & 92.81 & 74.31 & 58.91 & 
66.52 & 69.20 \\

\rowcolor{PTDRowBackground}
\textbf{PTD-PO(Ours)}
& \textbf{73.84} & \textbf{64.79} & \textbf{80.45} & 69.81 & \textbf{81.67} & \textbf{74.11}
& \textbf{41.20} & \textbf{92.88} & \textbf{76.15} & \textbf{60.32} & \textbf{67.64}
& \textbf{71.23} \\

\midrule

\multicolumn{13}{c}{\textbf{Qwen3-VL-8B-Thinking}} \\
\midrule
SFT 
& 55.97 & 59.09 & 67.72 & 57.81 & 70.47 & 62.21
& 34.14 & 76.06 & 66.74 & 46.09 & 55.76
& 59.34 \\

OPSD 
& 68.73 & 58.19 & 69.84 & 59.44 & 71.72 & 65.58
& 29.47 & 74.56 & 69.37 & 45.52 & 54.73
& 60.76 \\

GRPO 
& 70.75 & 60.17 & 77.19 & 67.97 & 80.17 & 71.25
& 42.83 & 88.56 & 73.98 & 57.38 & 65.69
& 68.78 \\

HDPO 
& 71.05 & \textbf{64.60} & 78.04 & 66.86 & 78.88 & 71.89 
& \textbf{45.87} & 90.06 & 74.55 & 59.40 & 67.47
& 69.92 \\

PAPO 
& 71.17 & 62.53 & 77.31 & 66.20 & 79.10 &  71.26
& 42.90 & 88.69 & 74.27 & 59.28 & 
66.29 & 69.05 \\

\rowcolor{PTDRowBackground}
\textbf{PTD-PO(Ours)}
& \textbf{76.21} & 64.52 & \textbf{79.81} & \textbf{71.39} & \textbf{82.31} & \textbf{74.85}
& 44.15 & \textbf{92.06} & \textbf{76.41} & \textbf{59.87} & \textbf{68.12}
& \textbf{71.86} \\

\bottomrule
\end{tabular}
}
\end{table*}
\vspace{-5pt}

As such, we have the compressed distributions:
\begin{equation}
\widetilde p_{i,t}
=
\left(
{p_{i,t}(v)}_{v\in \mathcal{S}_{i,t}},
p^{\mathrm{tail}}_{i,t}
\right),
\qquad
\widetilde q^h_{i,t}
=
\left(
{q^h_{i,t}(v)}_{v\in \mathcal{S}_{i,t}},
q^{h,\mathrm{tail}}_{i,t}
\right).
\end{equation}
After above approximations, our PTD divergence in \cref{eq:PTD} can be computed as
\begin{equation}
D_{\mathrm{PTD}}\left(q^h_{i,t},p_{i,t}\right)
\approx
D_{\mathrm{JSD}}\left(
\widetilde q^h_{i,t},
\widetilde p_{i,t}
\right).
\label{eq:PTD_JSD}
\end{equation}
It is worth mentioning that the tail bucket preserves omitted mass without renormalizing the retained tokens, so the objective in \cref{eq:PTD_JSD} is a mass-preserving coarse-grained lower bound on full JSD whose gap
contains only within-tail discrepancies (see Appendix~\ref{app:Top_K_JSD}), ensuring the computational reliability. Importantly, it reduces the storage required for distillation distributions from $\mathcal{O}(BTV)$ to
$\mathcal{O}(BTK)$ (with $|\mathcal{S}_{i,t}|\leq 2K$), while retaining high-probability
alignment signals
needed for stable privileged tutoring distillation.

\section{Experiments}
\subsection{Experimental Setup}

\paragraph{Training and evaluation.}
We use ViRL39K as the training corpus for multimodal reasoning post-training. Specifically, ViRL39K contains 38,870 verifiable vision-language question-answering examples and covers diverse reasoning scenarios, including mathematics, science, charts, diagrams, tables, documents, and spatial reasoning  \citep{lu2023mathvista, yue2024mmmu, xiao2024logicvista, zhang2024mathverse, johnson2017clevr, lu2021inter}. To support different paradigms, we construct three forms of auxiliary supervision for each example: our Way-revealing Structured Privileged Hints (WSPH), regular hints, and chain-of-thought answer trajectories with the final solution. For evaluation, we employ the PAPO \citep{wang2025perception} benchmark suite and its unified verifiable evaluation protocol, grouping tasks into general and vision-dependent multimodal reasoning and sampling eight rollouts per question to report average accuracy for each benchmark.

\paragraph{Implementation Details.}
For the main experiments, all model sizes are trained and evaluated with a maximum response length of 4096 tokens. For RL-based methods, including GRPO, self-distillation methods and PTD-PO, we use a rollout number of 8 during training. To reduce computational cost, ablation studies are conducted under a smaller but matched setting, with rollout number 5 and maximum response length 2048 tokens. More implementation details, hyperparameters, reward rules, and training configurations are provided in Appendix~\ref{app:more_impl}.

\subsection{Main Results}
\label{sec:main_results}


Table~\ref{tab:main_results} compares our PTD-PO with some state-of-the-arts spanning supervised fine-tuning, self-distillation, and multimodal RLVR by Qwen3-VL-Thinking models of different scales. PTD-PO consistently achieves the best overall performance on 2B, 4B, and 8B models, outperforming SFT, OPSD, and GRPO baseline \citep{zhao2026self, shao2024deepseekmath} on both general and vision-dependent multimodal reasoning tasks. Compared with HDPO \citep{ding2026hdpo} that directly conditions privileged self-distillation on complete CoT trajectories and ground-truth answers, our PTD-PO achieves clearly better results by transforming answer-derived privileged supervision into non-answer-revealing reasoning guidance rather than directly conditioning the tutor on the solution. PTD-PO also outperforms PAPO \citep{wang2025perception} substantially, which augments multimodal RLVR with perception-aware visual grounding objectives. These improvements suggest that reasoning-path tutoring provides gains beyond perception-oriented policy optimization. Overall, these results establish privileged tutoring distillation as a consistently effective complement to sparse outcome supervision across model scales. Training dynamics in Appendix~\ref{app:training_curves} further confirm its advantage.

\subsection{Compatibility and Broader Comparison}
Furthermore, we examine whether our PTD-PO can serve as a general auxiliary module across different RLVR objectives. As shown in Table~\ref{tab:ablation_rl_ptd} of Appendix~\ref{app:compatiable_rlvr}, we also integrate PTD with DAPO \citep{yu2026dapo} and GSPO \citep{zheng2025group} while retaining their original optimization objectives. The consistent improvements across these optimizers indicate that the proposed tutoring signal is not tied to a particular policy optimization rule, but can provide complementary corrective supervision for failed trajectories under different RLVR formulations. We also provide a broader comparison of PTD-PO with recent open-source multimodal reasoning methods, including VGPO \citep{wang2026vgpo} and CFPO \citep{yu2026cfpo}.

\subsection{Ablation Study}
\begin{figure*}[t]
    \centering
    \includegraphics[width=\textwidth]{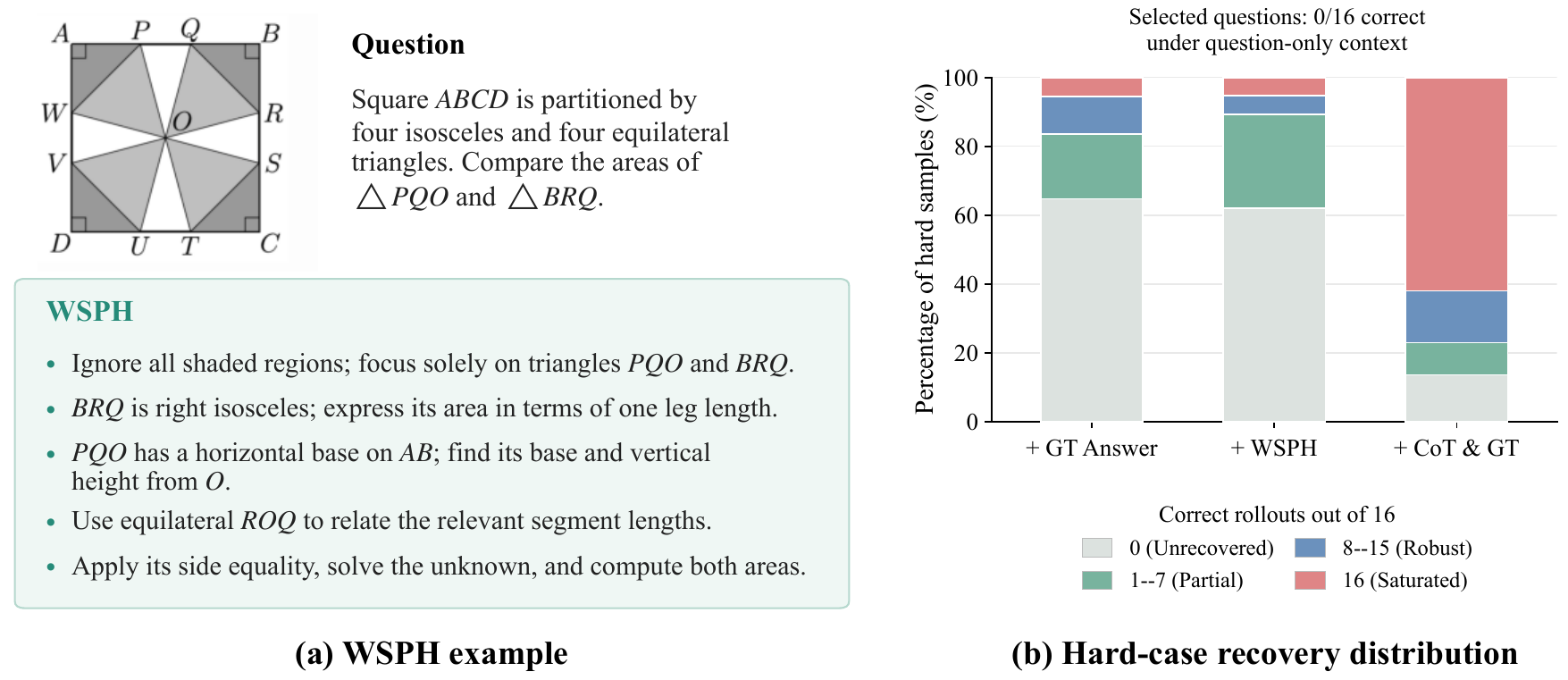}
    \caption{
    WSPH design and hard-case recovery on Qwen3-VL-4B-Thinking.
    (a) \textbf{WSPH example.} The hint identifies relevant subfigures and links their geometric constraints without revealing the answer.
    (b) \textbf{Hard-case recovery.} Recovery distributions across teacher contexts on questions with no successful question-only rollouts.
    }
    \label{fig:hint_ablation}
\end{figure*}

\paragraph{Effect of Way-revealing Tutoring Context.}

To examine the tutoring behavior induced by WSPH, we use Qwen3-VL-4B-Thinking and sample 16 rollouts for each of 384 examples under different teacher contexts. Figure~\ref{fig:hint_ablation}(b) focuses on hard questions with 0/16 correct rollouts under the question-only context. WSPH increases the fraction of these questions with 1--7 successful rollouts from 18.8\% under GT-answer prompting to 27.2\%. In contrast, CoT+GT yields 16/16 successes on 62.0\% of these questions, compared with 5.2\% under WSPH. These results distinguish partial recovery under WSPH from the predominantly all-pass behavior under full solution conditioning. Together with the distribution shift and response shortening in Figure~\ref{fig:teacher_context_shift}, they support using WSPH to provide corrective guidance with less reliance on solution-revealing contexts.

\begin{figure*}[t]
    \centering
    \includegraphics[width=\textwidth]{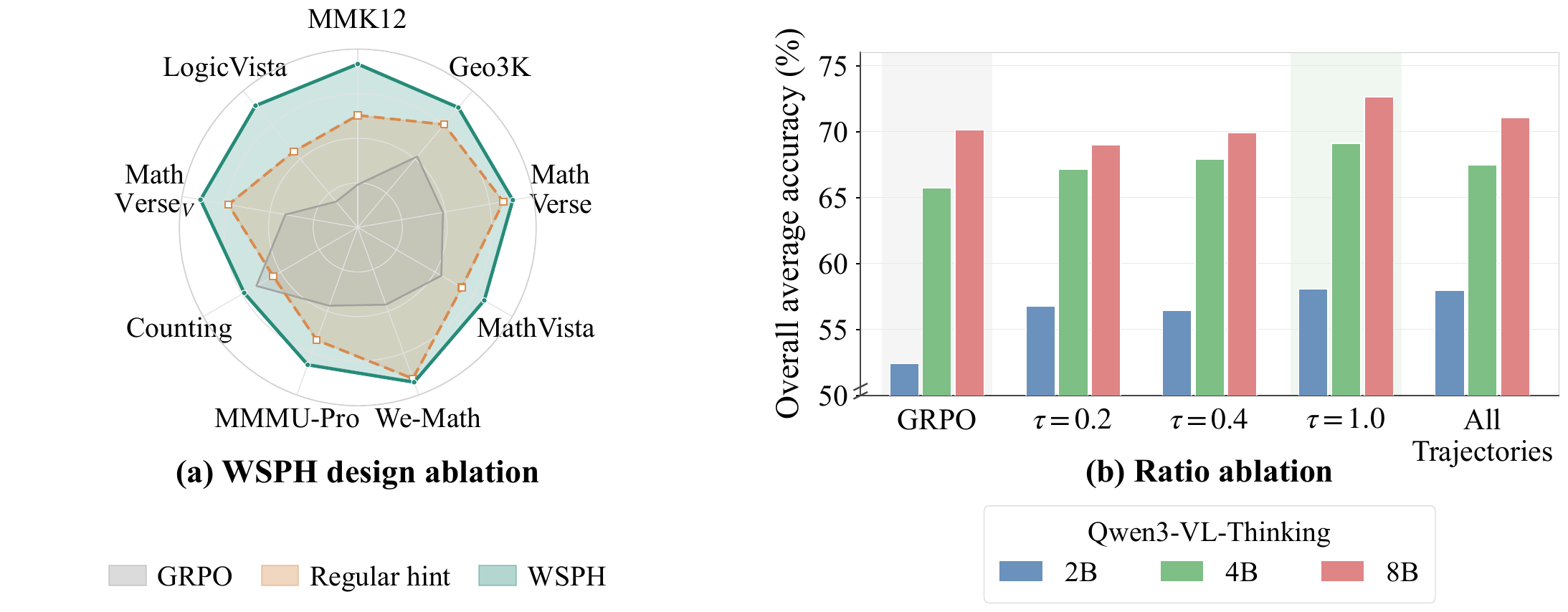}
    \vspace{-10pt}
    \caption{Ablations of PTD-PO. (a) Effect of hint design: GRPO, PTD with
  regular hints, and PTD with WSPH. (b) Effect of activation thresholds and
  trajectory selection on overall average accuracy.}
    \label{fig:ptd_ablation}
\end{figure*}


\paragraph{Effect of Way-revealing Structure.}
Figure~\ref{fig:ptd_ablation}(a) shows that even regular hints generally outperform GRPO, indicating that privileged reasoning direction alone provides useful supervision beyond sparse outcome rewards. WSPH consistently extends this improvement across all benchmarks, with a more pronounced advantage over regular hints on vision-dependent tasks such as LogicVista, Counting, and MathVerse$_V$. This separation indicates that the additional gain does not merely come from supplying more text: explicitly identifying relevant visual evidence and relations while filtering distractors helps the tutor correct perceptual selection and cross-modal binding errors that a regular hint leaves unresolved. The smaller separation on MathVerse and We-Math further supports this interpretation, since these tasks rely more on symbolic or textual reasoning once the problem has been parsed. Complete per-benchmark results are reported in Table~\ref{tab:ablation_PID-GRPO} of Appendix~\ref{app:detailed_ablation}.

\paragraph{Effect of PTD Activation Threshold.}
Finally, we assess effect of the trajectory selection strategy on our PTD-PO, where $\tau_{\mathrm{ptd}}$ denotes the group-level activation threshold and ``All Trajectories'' applies PTD to all rollouts regardless of correctness. Figure~\ref{fig:ptd_ablation}(b) shows that applying PTD to all failed trajectories with $\tau_{\mathrm{ptd}}=1.0$ achieves the highest overall average accuracy across the 2B, 4B, and 8B models. Smaller thresholds restrict PTD to more severely failed groups, reducing corrective supervision and yielding lower scores. Extending PTD to all trajectories also reduces performance, especially on larger models: correct rollouts already provide reliable reward-driven signals, and forcing them to match the WSPH-conditioned tutor may over-regularize the policy. These results support focusing PTD on failed trajectories, where answer-level rewards provide limited guidance.

\section{Conclusion}

In this paper, we presented PTD-PO, an RLVR framework that combines Way-revealing Structured Privileged Hints (WSPH) with failure-routed privileged tutoring distillation. WSPH provides where-to-look visual grounding and how-to-reason process guidance without exposing answer-identifying information or complete solution traces, while the student remains under the original question-only context. Failed rollouts are aligned with the WSPH-conditioned frozen reference model through a memory-efficient Top-$K$ JSD objective with tail compensation. Experiments on LVLMs from 2B to 8B show consistent improvements over RLVR and distillation baselines; regular hints already improve upon GRPO, while WSPH yields further gains, especially on vision-dependent tasks. These results clearly demonstrate that our idea of way-revealing privileged tutoring distillation can effectively complement sparse verifiable rewards by correcting perceptual grounding and reasoning-path errors without turning training into answer imitation.

\bibliographystyle{iclr2026_conference}
\bibliography{iclr2026_conference}
\clearpage
\appendix

\section*{\normalfont\scshape Appendix}
\phantomsection
\addcontentsline{toc}{section}{Appendix}

\vspace{1em}
{\normalfont\scshape Appendix Contents\par}
\vspace{0.8em}

\noindent\textbf{A\quad Related Work}\hfill \textbf{\pageref{app:related_work}}\\[0.8em]\

\textbf{B\quad More Implementation Details}\hfill \textbf{\pageref{app:more_impl}}\\[0.8em]

\textbf{C\quad Data Construction and Quality Control}\hfill \textbf{\pageref{app:data_construction}}\\[0.8em]

\hspace*{2em}C.1\quad WSPH Construction
\dotfill \pageref{app:hints_construction}\\[0.4em]

\hspace*{2em}C.2\quad Chain-of-Thought Answer Construction
\dotfill \pageref{app:cot_gt_construction}\\[0.4em]

\textbf{D\quad Proofs for Theoretical Analysis}\hfill \textbf{\pageref{app:proof-thm}}\\[0.4em]

\hspace*{2em}D.1\quad A Unified View of GRPO and Token-Level Distillation
\dotfill \pageref{app:GRPO_OPD_Theory}\\[0.4em]

\hspace*{2em}D.2\quad Why Jensen-Shannon Divergence Instead of Directional KL
\dotfill \pageref{app:Why_JSD}\\[0.4em]

\hspace*{2em}D.3\quad Approximation Property of Top-K JSD with Tail Compensation
\dotfill \pageref{app:Top_K_JSD}\\[0.4em]

\hspace*{2em}D.4\quad Optimization Rationale for Applying PTD to Failed Trajectories
\dotfill \pageref{app:Optimizatioin_rationale}\\[0.4em]

\textbf{E\quad Additional Experiments and Analysis}\hfill \textbf{\pageref{app:more_exp}}\\[0.4em]

\hspace*{2em}E.1\quad Compatibility With Other RLVR Optimizers
\dotfill \pageref{app:compatiable_rlvr}\\[0.4em]

\hspace*{2em}E.2\quad Sensitivity Analysis on the Top-$K$ Support Size
\dotfill \pageref{app:abl_Top_K}\\[0.4em]

\hspace*{2em}E.3\quad Training Dynamics and Efficiency Analysis
\dotfill \pageref{app:training_curves}\\[0.4em]

\hspace*{2em}E.4\quad Detailed Ablation Results
\dotfill \pageref{app:detailed_ablation}\\[0.4em]

\textbf{F\quad Case Study}\hfill \textbf{\pageref{app:case_study}}\\[0.4em]

\textbf{G\quad Limitations and Future Work}\hfill \textbf{\pageref{app:limitations}}\\[0.4em]

\clearpage

\section{Related Work}
\label{app:related_work}

\paragraph{RLVR for Multimodal Reasoning.}
Reinforcement Learning with Verifiable Rewards (RLVR) uses automatically checkable outcomes rather than learned RLHF rewards~\citep{schulman2017proximal,ouyang2022training,askell2021general}. Text-based advances include DeepSeekMath, DeepSeek-R1, Kimi k1.5, and improved policy optimizers~\citep{shao2024deepseekmath,guo2025deepseek,team2025kimi,yu2026dapo,zheng2025group,zhao2025geometric}. Multimodal extensions such as Vision-R1, R1-VL, MM-Eureka, OpenVLThinker, and Reason-RFT combine cold-start data, rule-based rewards, and GRPO-style optimization~\citep{huang2025vision,zhou2503r1,meng2025mm,deng2025openvlthinker,tan2025reason}. ThinkLite-VL uses MCTS-guided sample selection~\citep{wang2025thinklite}, VL-Rethinker combines selective sample replay with forced rethinking~\citep{wang2025vlrethinker}, and NoisyRollout augments images to enhance exploration~\citep{liu2025noisyrollout}. For visual grounding, PAPO adds an implicit perception loss~\citep{wang2025perception}, and VPPO-RL reweights trajectory advantages and selects policy-update tokens using visual dependency~\citep{huang2025spotlight}. Other methods explore contrastive perception learning and visual re-focusing~\citep{lin2025accelerating,yang2026look}. VGPO combines visual attention compensation with token- and trajectory-level advantage reweighting~\citep{wang2026vgpo}, while CFPO regularizes predictions through counterfactual suppression of critical visual cues~\citep{yu2026cfpo}. In contrast, PTD-PO densely corrects failed trajectories through distillation from a frozen self-teacher conditioned on WSPH, while preserving the student's question-only context.

\paragraph{On-Policy and Self-Distillation for Dense Supervision.}
Another closely related direction is on-policy distillation (OPD), which supervises student-generated trajectories with dense teacher distributions, combining the distributional relevance of on-policy learning with token-level feedback from knowledge distillation~\citep{bai2025qwen3, ye2026policy, ye2025black}. Compared with off-policy distillation on fixed teacher traces, OPD can better expose the teacher to student-side errors, but it still typically depends on an external stronger teacher and incurs nontrivial rollout-time supervision cost~\citep{li2026rethinking}. Recent self-distillation methods further reduce this dependency by letting the same model serve as both student and teacher under different contexts, e.g., conditioning the teacher on ground-truth solutions, rich feedback, successful sibling rollouts, or other privileged signals to provide dense token-level correction~\citep{zhao2026self, hubotter2026reinforcement,li2026unifying}. Vision-OPD extends this idea to multimodal fine-grained perception by using a crop-conditioned policy as a privileged teacher to supervise a full-image policy on on-policy rollouts~\citep{yuan2026vision}. More relevant to our work, HDPO introduces privileged self-distillation into RL by directly using ground-truth answers to construct a self-teacher for cliff prompts where standard RL yields vanishing gradients~\citep{ding2026hdpo}. However, HDPO focuses on text-only mathematical reasoning and relies on answer-revealing privileged information, while Vision-OPD focuses on crop-based regional perception; in contrast, our method targets multimodal reasoning and uses WSPH to provide non-answer-revealing visual-grounding and reasoning guidance for correcting failed trajectories.

\section{More Implementation Details}
\label{app:more_impl}

\paragraph{Overall Setup.}
All experiments are implemented with PyTorch and conducted on Qwen3-VL-Thinking models with different parameter scales. For all training methods, we use ViRL39K as the training corpus and train each model for two epochs. The vision encoder is kept consistent with the original model configuration, and all RL-based methods are trained under the same data mixture, rollout setting, response length budget, and evaluation protocol to ensure fair comparison. For PTD-PO, WSPH is used only to condition the frozen reference model for teacher-side distribution generation and is never included in the student rollout or evaluation context.

\paragraph{Training Details.}
For the main experiments, the maximum response length is set to 4096 tokens during both training and evaluation. Ablation studies use a maximum response length of 2048 tokens and five training rollouts per prompt. SFT is optimized with standard next-token prediction on the constructed solution trajectories. For RL-based methods, each prompt is sampled under the original question-only context, and the policy generates multiple rollouts for verifiable reward optimization. The reward is composed of a format reward and an accuracy reward, with weights of $0.1$ and $0.9$, respectively, where the accuracy reward is computed by rule-based answer matching whenever possible. GRPO and PTD-PO share the same rollout configuration and reward design. OPSD follows its original training settings, with no additional modification except using the same training data, response length budget, and evaluation protocol for fair comparison. For PTD-PO, teacher-side logits are computed only for failed trajectories activated by the PTD criterion and are further compressed by the Top-$K$ JSD objective. All experiments are conducted on 8 or 16 NVIDIA H100 GPUs depending on the model size.

\paragraph{Evaluation Details.}
All experiments follow a train-inference consistent setting. During both student rollout training and benchmark evaluation, the policy model is conditioned only on the original question context, without access to privileged hints, spatial guidance, intermediate textual hints, or ground-truth answers. We use the same response length budget and answer extraction protocol for all compared methods. The final answer is extracted from the model response and evaluated by dataset-specific rule-based matching whenever possible, and free-form instances are filtered following the main evaluation protocol to avoid LLM-as-a-judge.

\begin{table}[htp]
\centering
\caption{Key hyperparameters for training and evaluation.}
\label{tab:implementation_details}
\renewcommand{\arraystretch}{1.15}
\setlength{\tabcolsep}{4.5mm}
\begin{tabular}{ll}
\toprule
\textbf{Hyperparameter} & \textbf{Value} \\
\midrule
\multicolumn{2}{l}{\textit{General Training}} \\
Base Models & Qwen3-VL-2B/4B/8B-Thinking \\
Training Data & ViRL39K \\
Training Epochs & 2 \\
Optimizer & AdamW \\
Learning Rate & \textit{1e-6} \\
Weight Decay & \textit{1e-2} \\
LR Schedule & \textit{Constant} \\
Warmup Ratio & \textit{0} \\
Precision & \textit{bf16} \\
Freeze Vision Tower & \textit{False} \\
\midrule
\multicolumn{2}{l}{\textit{RLVR Training}} \\
Global Batch Size & 128 \\
Rollout Batch Size & 384 \\
Rollout Top-p & 0.99 \\
Rollout Number & 8 \\
Ablation Rollout Number & 5 \\
Max Response Length & 4096 \\
Ablation Max Response Length & 2048 \\
Reward Type & Format + Accuracy \\
Format Reward Weight & 0.1 \\
Accuracy Reward Weight & 0.9 \\
Accuracy Reward & Binary accuracy $(1/0)$ \\
KL Penalty & 1e-2 \\
\midrule
\multicolumn{2}{l}{\textit{PTD-PO Specific}} \\
$\lambda_{ptd}$ &  \{5e-2, 5e-1\}\\
$\tau_{\mathrm{ptd}}$ & 1.0 \\
PTD Trajectory Selection & Failed trajectories only \\
Divergence Objective & Top-$K$ JSD with tail compensation \\
Top-$K$ Size & 100 \\
Teacher Model & Frozen reference model \\
\midrule
\multicolumn{2}{l}{\textit{Evaluation Generation}} \\
Evaluation Context & Question only \\
Temperature & 1.0 \\
Top-$p$ & 1.0 \\
Max New Tokens & 4096 \\
\bottomrule
\end{tabular}
\end{table}

\section{Data Construction and Quality Control}
\label{app:data_construction}

\subsection{WSPH Construction}
\label{app:hints_construction}

\begin{figure}[t]
    \centering
    \includegraphics[width=0.96\textwidth]{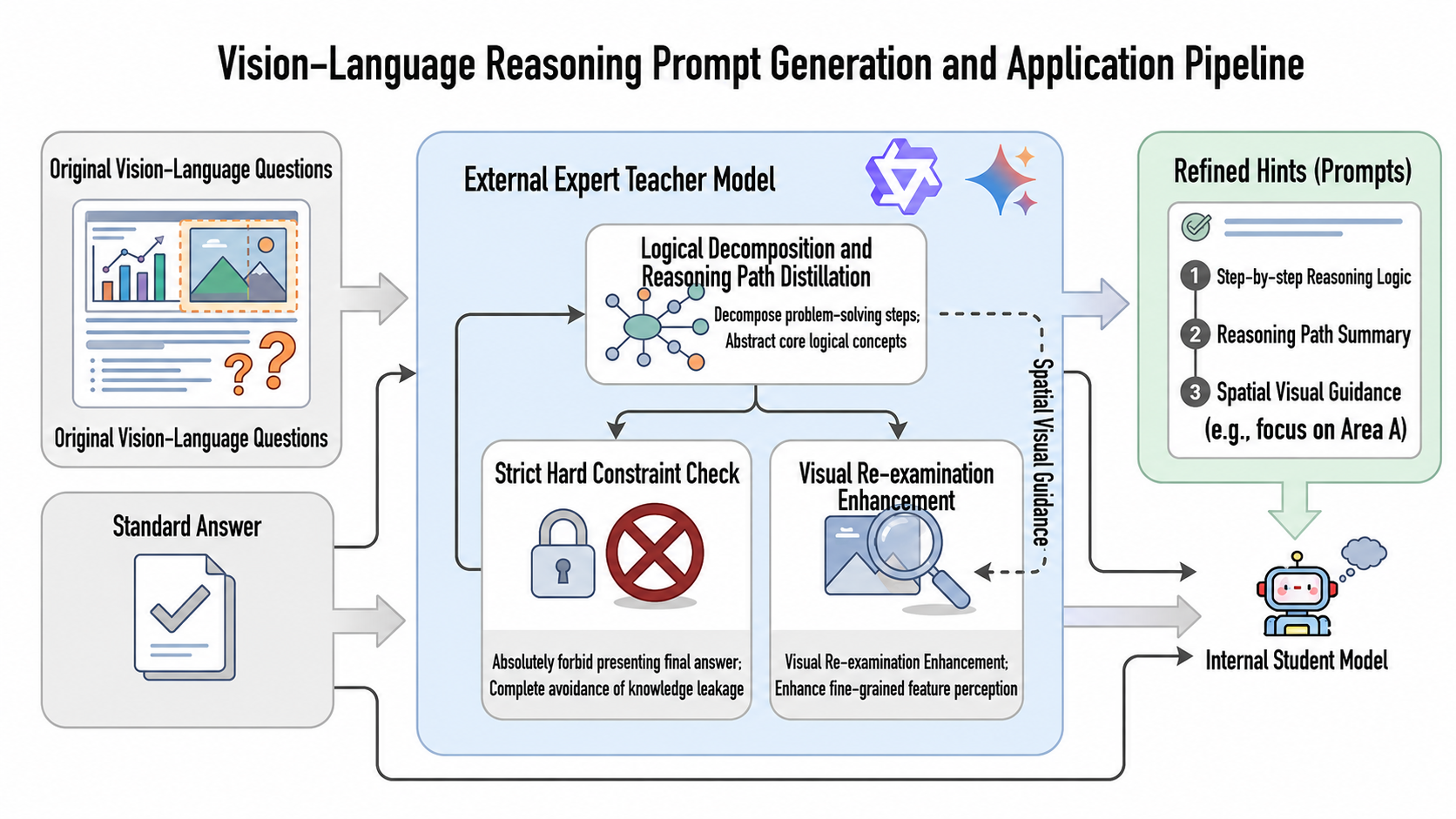}
    \caption{Way-revealing structured privileged hint (WSPH) construction pipeline.}
    \label{fig:hint_generation_pipeline}
\end{figure}

\begin{figure}[p]
    \centering
    \includegraphics[width=0.92\textwidth,height=0.82\textheight,keepaspectratio]{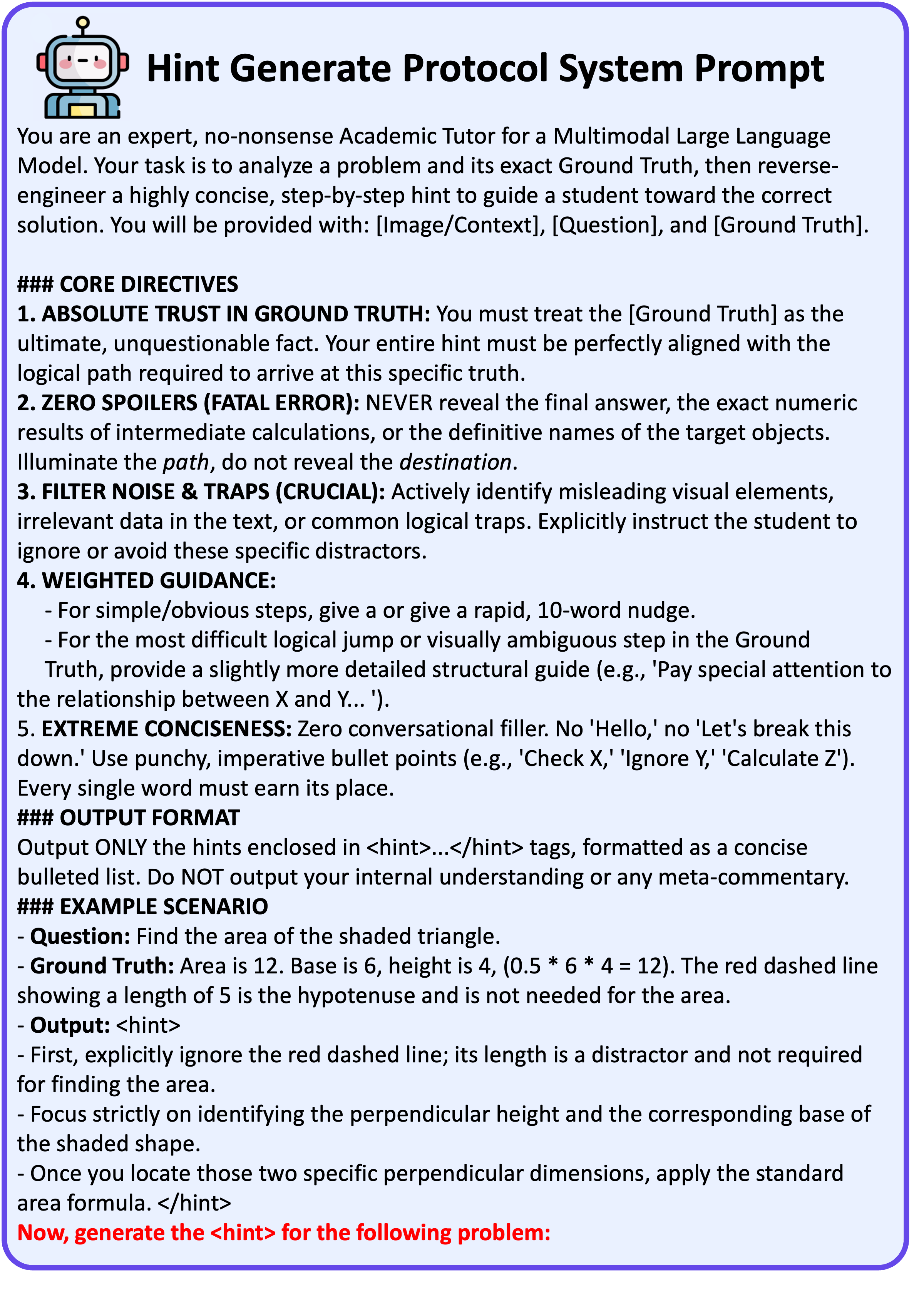}
    \caption{Prompt used to generate WSPH.}
    \label{fig:hint_system_prompt}
\end{figure}

As shown in Figure~\ref{fig:hint_generation_pipeline}, we construct Way-revealing Structured Privileged Hints (WSPH) offline from the image-context, the question, and the verified ground-truth answer through two stages: way-revealing hint generation and hint refinement. The external model is used only as a hint generator before PTD training, and is not involved in rollout, distillation, or evaluation. For each instance, we provide the generator with $(I,q,a^\star)$ and the system prompt in Figure~\ref{fig:hint_system_prompt}, and ask it to reverse-engineer a concise hint that points to the necessary visual evidence and reasoning route without producing a complete solution trace.

In the way-revealing hint generation stage, we mainly use Qwen3-VL-235B-A22B-Thinking to generate WSPH candidates. For particularly difficult cases where this model fails to produce a complete hint within the maximum generation budget of 16,384 tokens, we fall back to Gemini-3.0-Pro to generate the hint. The same system prompt is used for both generators to ensure a consistent annotation style, combining where-to-look visual grounding with how-to-reason guidance. Specifically, the prompt enforces three requirements: the hint should be aligned with the verified reasoning path, should not reveal the final answer or exact intermediate results, and should explicitly suppress irrelevant visual elements or misleading textual cues. It also encourages concise imperative instructions, with slightly more detailed guidance only for visually ambiguous or logically difficult steps.

In the hint refinement stage, we apply a lightweight quality-control pipeline to remove or regenerate invalid or potentially answer-revealing hints. An independent verifier model automatically inspects each hint for explicit final answers, decisive intermediate numerical results, strongly answer-identifying cues, missing key visual evidence, or degeneration into a full chain-of-thought solution. Hints that violate any of these criteria are discarded or regenerated, while the remaining validated WSPH annotations condition the frozen reference model to provide token-level reasoning-path supervision through PTD.

\subsection{Chain-of-Thought Answer Construction}
\label{app:cot_gt_construction}

We use Qwen3-VL-235B-A22B-Thinking and Gemini-3.0-Pro as teacher models to construct SFT chain-of-thought answers for ViRL39K. For each training instance, the input image, question, and thinking instruction are provided to the teacher, requiring the response to contain a reasoning process enclosed by \texttt{<think>} and \texttt{</think>} and a final answer enclosed by \texttt{\textbackslash boxed\{\}}. We sample multiple responses for each question and retain only candidates whose extracted boxed answer matches the ground-truth answer under the rule-based accuracy checker. Among all correct candidates, we keep the shortest one to reduce unnecessary verbosity. The accepted samples are saved in ShareGPT-style JSONL format, together with the corresponding image path.

For questions not solved in the initial generation pass, we apply an additional repair stage. We first re-sample missing instances with a larger generation budget. If they are still unresolved, we optionally use a reverse-thinking pass, where the teacher is given the ground-truth answer only during data construction and is asked to back-construct a coherent reasoning chain. The stored SFT example still contains only the original question and standard thinking instruction, without exposing the answer hint in the user message. All repaired responses are again verified by the same rule-based answer checker, and samples that fail format extraction or answer matching are discarded.

\section{Proofs for Theoretical Analysis}
\label{app:proof-thm}
\subsection{A Unified View of GRPO and Token-Level Distillation}
\label{app:GRPO_OPD_Theory}

We provide a distributional view of GRPO and token-level distillation. 
Let $s=(x,y_{<t})$ denote a token-generation state and let $a\in\mathcal{V}$ denote the next token. 
For a fixed state $s$, consider the following regularized policy improvement problem:
\begin{equation}
\pi^{+}(\cdot|s)
=
\arg\max_{\pi(\cdot|s)}
\left[
\mathbb{E}_{a\sim \pi(\cdot|s)} A(s,a)
-
\tau D_{\mathrm{KL}}
\left(
\pi(\cdot|s)
\|
\pi_{\mathrm{old}}(\cdot|s)
\right)
\right],
\label{eq:app_regularized_policy_improvement}
\end{equation}
where $A(s,a)$ is the token-level advantage, $\pi_{\mathrm{old}}$ is the rollout policy, and $\tau>0$ is a temperature coefficient.

\paragraph{Lemma D.1.}
The solution of Eq.~\eqref{eq:app_regularized_policy_improvement} is
\begin{equation}
\pi^{+}(a|s)
=
\frac{
\pi_{\mathrm{old}}(a|s)\exp(A(s,a)/\tau)
}{
Z(s)
},
\quad
Z(s)=
\sum_{a'\in\mathcal{V}}
\pi_{\mathrm{old}}(a'|s)\exp(A(s,a')/\tau).
\label{eq:app_improved_policy}
\end{equation}

\paragraph{Proof.}
The Lagrangian of Eq.~\eqref{eq:app_regularized_policy_improvement} is
\begin{equation}
\mathcal{L}(\pi,\eta)
=
\sum_{a\in\mathcal{V}}\pi(a|s)A(s,a)
-
\tau
\sum_{a\in\mathcal{V}}
\pi(a|s)
\log
\frac{\pi(a|s)}{\pi_{\mathrm{old}}(a|s)}
+
\eta
\left(
\sum_{a\in\mathcal{V}}\pi(a|s)-1
\right).
\end{equation}
Taking the derivative with respect to $\pi(a|s)$ gives
\begin{equation}
A(s,a)
-
\tau
\left[
\log
\frac{\pi(a|s)}{\pi_{\mathrm{old}}(a|s)}
+1
\right]
+\eta
=0.
\end{equation}
Thus,
\begin{equation}
\pi(a|s)
=
\pi_{\mathrm{old}}(a|s)
\exp
\left(
\frac{A(s,a)}{\tau}
\right)
\exp
\left(
\frac{\eta-\tau}{\tau}
\right).
\end{equation}
Normalizing over $\mathcal{V}$ yields Eq.~\eqref{eq:app_improved_policy}.
\hfill $\square$

Eq.~\eqref{eq:app_improved_policy} shows that the advantage function induces a reward-improved target distribution. Equivalently, from Eq.~\eqref{eq:app_improved_policy},
\begin{equation}
\log \pi^{+}(a|s)
=
\log \pi_{\mathrm{old}}(a|s)
+
\frac{1}{\tau}A(s,a)
-
\log Z(s).
\end{equation}
For any policy $\pi$, we have
\begin{align}
D_{\mathrm{KL}}
\left(
\pi(\cdot|s)
\|
\pi^{+}(\cdot|s)
\right)
&=
\sum_{a\in\mathcal{V}}
\pi(a|s)
\log
\frac{\pi(a|s)}{\pi^{+}(a|s)}
\nonumber\\
&=
\sum_{a\in\mathcal{V}}
\pi(a|s)
\log
\frac{\pi(a|s)}{\pi_{\mathrm{old}}(a|s)}
-
\frac{1}{\tau}
\mathbb{E}_{a\sim\pi(\cdot|s)}A(s,a)
+
\log Z(s)
\nonumber\\
&=
D_{\mathrm{KL}}
\left(
\pi(\cdot|s)
\|
\pi_{\mathrm{old}}(\cdot|s)
\right)
-
\frac{1}{\tau}
\mathbb{E}_{a\sim\pi(\cdot|s)}A(s,a)
+
\log Z(s).
\label{eq:app_kl_equivalence}
\end{align}
Since $\log Z(s)$ is independent of $\pi$, maximizing Eq.~\eqref{eq:app_regularized_policy_improvement} is equivalent to minimizing
\begin{equation}
D_{\mathrm{KL}}
\left(
\pi(\cdot|s)
\|
\pi^{+}(\cdot|s)
\right).
\label{eq:app_grpo_as_target_matching}
\end{equation}

Therefore, GRPO can be interpreted as implicit distributional policy improvement. 
Its scalar advantage defines an implicit target distribution $\pi^{+}$, and the policy update moves $\pi_{\theta}$ toward this reward-improved distribution.

In practical GRPO, the token-level advantage is estimated from group-relative outcome rewards. 
For a sampled response $y_i$, the same group-level advantage $A_i$ is assigned to all valid tokens in the response:
\begin{equation}
A(s_{i,t},y_{i,t}) \approx A_i,
\quad
s_{i,t}=(x,y_{i,<t}).
\end{equation}
The clipped GRPO surrogate can thus be viewed as a sample-based approximation to the distributional improvement in Eq.~\eqref{eq:app_grpo_as_target_matching}, with clipping used to constrain the update around $\pi_{\theta_{\mathrm{old}}}$.

Token-level distillation provides an explicit distributional target. 
Given a teacher distribution $q(\cdot|s)$ and a student distribution $p_{\theta}(\cdot|s)$, the standard distillation objective is
\begin{equation}
\mathcal{L}_{\mathrm{OPD}}
=
\mathbb{E}_{s}
\left[
D
\left(
q(\cdot|s),
p_{\theta}(\cdot|s)
\right)
\right],
\label{eq:app_td_general}
\end{equation}
where $D(\cdot,\cdot)$ can be instantiated as KL divergence, Jensen-Shannon divergence, or other distributional discrepancies. 
When $D=D_{\mathrm{KL}}(q\|p_{\theta})$, the objective becomes
\begin{equation}
\mathcal{L}_{\mathrm{OPD}}
=
\mathbb{E}_{s}
\left[
\sum_{a\in\mathcal{V}}
q(a|s)
\log
\frac{q(a|s)}{p_{\theta}(a|s)}
\right].
\label{eq:app_forward_kl_td}
\end{equation}
With $q$ fixed, minimizing Eq.~\eqref{eq:app_forward_kl_td} is equivalent to minimizing the cross-entropy
\begin{equation}
-
\mathbb{E}_{s}
\left[
\sum_{a\in\mathcal{V}}
q(a|s)
\log p_{\theta}(a|s)
\right],
\end{equation}
which directly matches the student distribution to the teacher distribution at each token position.

GRPO and token-level distillation can therefore be written in a unified form:
\begin{equation}
\min_{\theta}
\;
\mathbb{E}_{s}
\left[
\underbrace{
D_{\mathrm{KL}}
\left(
p_{\theta}(\cdot|s)
\|
\pi^{+}(\cdot|s)
\right)
}_{\text{implicit reward-induced target}}
+
\lambda
\underbrace{
D
\left(
q(\cdot|s),
p_{\theta}(\cdot|s)
\right)
}_{\text{explicit teacher-induced target}}
\right].
\label{eq:app_unified_objective}
\end{equation}
The first term corresponds to reward-driven policy improvement, where the target distribution is induced by group-relative advantages. 
The second term corresponds to token-level distillation, where the target distribution is explicitly provided by a teacher.

In PTD-PO, the explicit teacher target is provided by the WSPH-conditioned frozen reference model, where $h$ denotes the WSPH constructed in Appendix~\ref{app:hints_construction}:
\begin{equation}
q^{h}_{i,t}
=
\pi_{\mathrm{ref}}
\left(
\cdot
\mid
x^{h},y_{i,<t}
\right),
\quad
x^{h}=[x;h],
\label{eq:app_ptd_teacher}
\end{equation}
while the student remains conditioned on the original question-only context:
\begin{equation}
p_{i,t}
=
\pi_{\theta}
\left(
\cdot
\mid
x,y_{i,<t}
\right).
\label{eq:app_ptd_student}
\end{equation}
Thus, PTD-PO does not replace reward optimization with imitation learning. 
It augments the implicit reward-induced target of GRPO with an explicit token-level corrective target on selected trajectories:
\begin{equation}
\mathcal{L}_{\mathrm{PTD\mbox{-}PO}}
=
L_{\mathrm{GRPO}}(\theta)
+
\lambda_{\mathrm{PTD}}
\mathcal{L}_{\mathrm{PTD}}.
\label{eq:app_ptd_grpo_final}
\end{equation}

This establishes the theoretical role of PTD: GRPO provides an on-policy reward-improvement direction, while PTD provides a token-level distributional correction when the reward-induced target is under-informative.

\subsection{Why Jensen-Shannon Divergence Instead of Directional KL}
\label{app:Why_JSD}
We analyze the optimization behavior of directional KL objectives and Jensen-Shannon divergence under the asymmetric context setting of PTD-PO. 
Let
\begin{equation}
q(\cdot|s)=\pi_{\mathrm{ref}}(\cdot|x^h,y_{<t}),
\quad
p_{\theta}(\cdot|s)=\pi_{\theta}(\cdot|x,y_{<t}),
\end{equation}
where the teacher distribution $q$ is conditioned on the WSPH-augmented context $x^h=[x;h]$, with $h$ denoting WSPH, while the student distribution $p_{\theta}$ is conditioned on the original question-only context $x$.

\begin{figure}[t]
    \centering
    \includegraphics[width=0.98\textwidth]{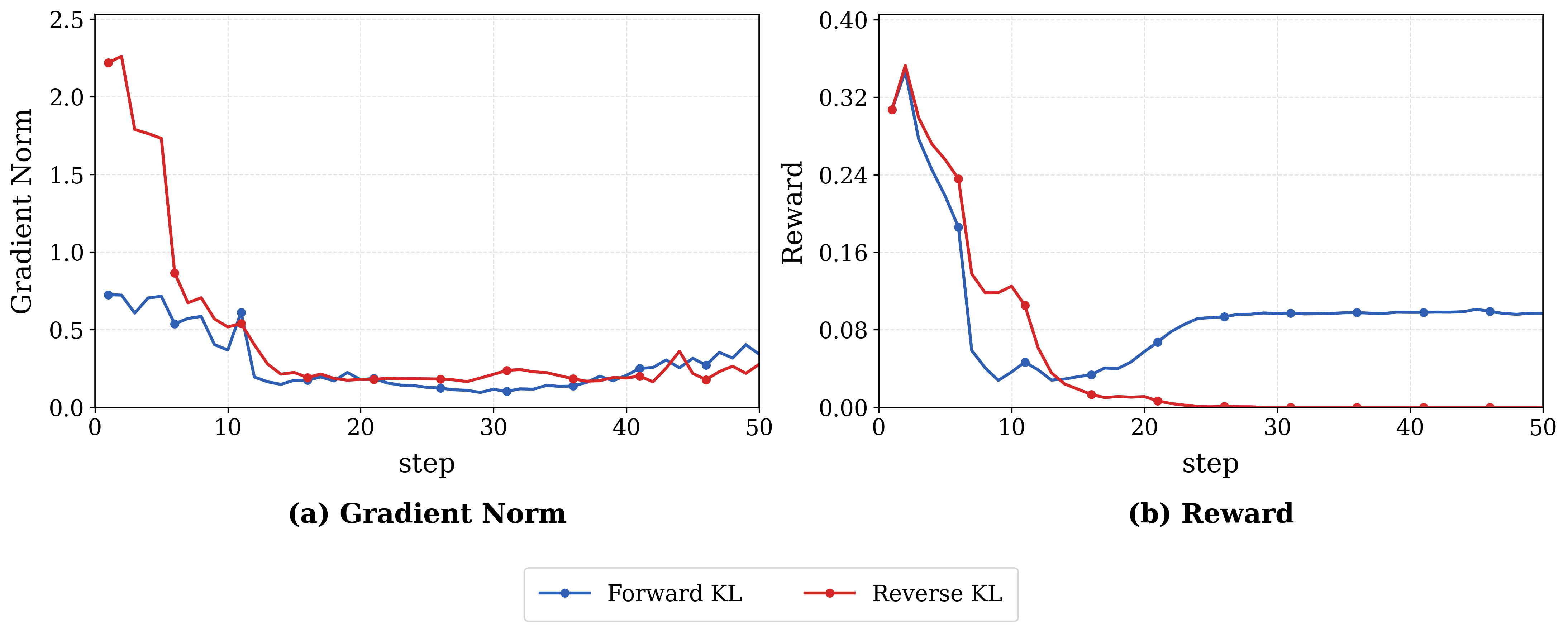}
    \caption{
    Comparison between forward KL and reverse KL during PTD-style token-level distillation.
    Forward KL produces a more moderate gradient norm but yields weaker reward improvement, while reverse KL introduces substantially larger early-stage gradients and quickly collapses the reward signal.
    }
    \label{fig:kl_gradient_reward}
\end{figure}

\paragraph{Forward KL.}
The forward KL objective is
\begin{equation}
D_{\mathrm{KL}}(q\|p_{\theta})
=
\sum_{v\in\mathcal{V}}
q(v)
\log
\frac{q(v)}{p_{\theta}(v)}.
\end{equation}
Let $z_v$ denote the student logit and $p_{\theta}(v)=\mathrm{softmax}(z)_v$. 
Since $q$ is fixed, the gradient of forward KL with respect to the student logit is
\begin{equation}
\frac{\partial D_{\mathrm{KL}}(q\|p_{\theta})}{\partial z_v}
=
p_{\theta}(v)-q(v).
\label{eq:app_forward_kl_gradient}
\end{equation}
Thus, forward KL directly pulls the student distribution toward the teacher distribution at every token position. 
When $q$ is produced under a hint-augmented context, the difference $q(v)-p_{\theta}(v)$ contains not only corrective reasoning information, but also context-induced distribution shift. 
Therefore, forward KL may force the question-only student to imitate probability mass that is specific to the privileged context, rather than learning a stable reward-aligned correction.

\paragraph{Reverse KL.}
The reverse KL objective is
\begin{equation}
D_{\mathrm{KL}}(p_{\theta}\|q)
=
\sum_{v\in\mathcal{V}}
p_{\theta}(v)
\log
\frac{p_{\theta}(v)}{q(v)}.
\end{equation}
Its gradient with respect to the student logit is
\begin{equation}
\frac{\partial D_{\mathrm{KL}}(p_{\theta}\|q)}{\partial z_v}
=
p_{\theta}(v)
\left[
\log\frac{p_{\theta}(v)}{q(v)}
-
\mathbb{E}_{u\sim p_{\theta}}
\left(
\log\frac{p_{\theta}(u)}{q(u)}
\right)
\right].
\label{eq:app_reverse_kl_gradient}
\end{equation}
Reverse KL penalizes assigning probability mass to tokens with small teacher probability. 
When the hint-conditioned teacher becomes sharper than the question-only student, tokens that remain plausible under the student but receive low probability under the teacher can induce large logarithmic penalties. 
This produces a mode-seeking effect and can suppress the exploratory distribution maintained by RLVR.

Figure~\ref{fig:kl_gradient_reward} empirically illustrates this behavior. 
Reverse KL yields much larger gradient norms in the early optimization stage, while its reward quickly degenerates. 
Forward KL is more stable in gradient magnitude, but its reward improvement is weaker. 
These results indicate that directional KL objectives are not well matched to the asymmetric teacher-student contexts in PTD-PO.

\paragraph{Jensen-Shannon divergence.}
PTD-PO instead uses Jensen-Shannon divergence:
\begin{equation}
D_{\mathrm{JSD}}(q,p_{\theta})
=
\frac{1}{2}
D_{\mathrm{KL}}(q\|m)
+
\frac{1}{2}
D_{\mathrm{KL}}(p_{\theta}\|m),
\quad
m=\frac{1}{2}(q+p_{\theta}).
\label{eq:app_jsd}
\end{equation}
Different from directional KL, JSD aligns both distributions through the mixture distribution $m$. 
For each token $v$, the denominator in the logarithmic ratio is not the opposite distribution alone, but the smoothed mixture
\begin{equation}
m(v)=\frac{1}{2}\left(q(v)+p_{\theta}(v)\right).
\end{equation}
This prevents either distribution from acting as an overly strict one-sided target.

The derivative of JSD with respect to the student probability has a simple form:
\begin{equation}
\frac{\partial D_{\mathrm{JSD}}(q,p_{\theta})}{\partial p_{\theta}(v)}
=
\frac{1}{2}
\log
\frac{p_{\theta}(v)}{m(v)}
=
\frac{1}{2}
\log
\frac{2p_{\theta}(v)}{p_{\theta}(v)+q(v)}.
\label{eq:app_jsd_prob_gradient}
\end{equation}
Accordingly, the gradient with respect to the student logit is
\begin{equation}
\frac{\partial D_{\mathrm{JSD}}(q,p_{\theta})}{\partial z_v}
=
\frac{1}{2}p_{\theta}(v)
\left[
\log
\frac{p_{\theta}(v)}{m(v)}
-
\mathbb{E}_{u\sim p_{\theta}}
\left(
\log
\frac{p_{\theta}(u)}{m(u)}
\right)
\right].
\label{eq:app_jsd_logit_gradient}
\end{equation}
Compared with Eq.~\eqref{eq:app_reverse_kl_gradient}, JSD replaces the teacher denominator $q(v)$ with the mixture denominator $m(v)$. 
Thus, even when $q(v)$ is very small, the penalty is softened by the student probability $p_{\theta}(v)$ through $m(v)$. 
This reduces the risk of extremely aggressive gradients caused by context-induced teacher sharpness.

\paragraph{Boundedness.}
JSD is bounded:
\begin{equation}
0
\le
D_{\mathrm{JSD}}(q,p_{\theta})
\le
\log 2.
\label{eq:app_jsd_bound}
\end{equation}
The lower bound follows from the non-negativity of KL divergence. 
For the upper bound, since
\begin{equation}
m(v)=\frac{q(v)+p_{\theta}(v)}{2}
\ge
\frac{q(v)}{2},
\quad
m(v)
\ge
\frac{p_{\theta}(v)}{2},
\end{equation}
we have
\begin{equation}
D_{\mathrm{KL}}(q\|m)
=
\sum_{v}q(v)\log\frac{q(v)}{m(v)}
\le
\sum_{v}q(v)\log 2
=
\log 2,
\end{equation}
and similarly,
\begin{equation}
D_{\mathrm{KL}}(p_{\theta}\|m)
\le
\log 2.
\end{equation}
Substituting these two inequalities into Eq.~\eqref{eq:app_jsd} gives Eq.~\eqref{eq:app_jsd_bound}.

Therefore, JSD provides a symmetric and bounded distributional discrepancy. 
In PTD-PO, this property is important because the teacher and student are intentionally conditioned on different contexts:
\begin{equation}
q(\cdot|s)
=
\pi_{\mathrm{ref}}(\cdot|x^h,y_{<t}),
\quad
p_{\theta}(\cdot|s)
=
\pi_{\theta}(\cdot|x,y_{<t}).
\end{equation}
The objective should transfer corrective reasoning information from the hint-conditioned teacher, but should not force the student to fully reproduce the privileged-context distribution. 
JSD satisfies this requirement by using a softened mixture target and by avoiding the one-sided optimization bias of directional KL objectives.

\subsection{Approximation Property of Top-K JSD with Tail Compensation}
\label{app:Top_K_JSD}

We analyze the approximation property of the proposed Top-K JSD with tail compensation. 
For simplicity, we omit the token position index $(i,t)$ and write the student and teacher distributions as $p$ and $q$ over the vocabulary $\mathcal{V}$. 
Let $S\subset\mathcal{V}$ be the compact support used by Top-K JSD:
\begin{equation}
S
=
\operatorname{TopK}(p)
\cup
\operatorname{TopK}(q).
\end{equation}
The remaining vocabulary set is denoted by
\begin{equation}
\bar{S}
=
\mathcal{V}\setminus S.
\end{equation}
The tail masses of $p$ and $q$ are
\begin{equation}
p_{\mathrm{tail}}
=
\sum_{v\in\bar{S}}p(v)
=
1-\sum_{v\in S}p(v),
\quad
q_{\mathrm{tail}}
=
\sum_{v\in\bar{S}}q(v)
=
1-\sum_{v\in S}q(v).
\end{equation}
The compressed distributions are defined as
\begin{equation}
\tilde{p}
=
\left(
\{p(v)\}_{v\in S},
p_{\mathrm{tail}}
\right),
\quad
\tilde{q}
=
\left(
\{q(v)\}_{v\in S},
q_{\mathrm{tail}}
\right).
\end{equation}

\paragraph{Validity of the compressed distributions.}
The compressed distributions remain valid probability distributions:
\begin{equation}
\sum_{v\in S}\tilde{p}(v)+\tilde{p}_{\mathrm{tail}}
=
\sum_{v\in S}p(v)
+
\left(
1-\sum_{v\in S}p(v)
\right)
=
1,
\end{equation}
and similarly,
\begin{equation}
\sum_{v\in S}\tilde{q}(v)+\tilde{q}_{\mathrm{tail}}
=
1.
\end{equation}
Thus, Top-K JSD is computed on a normalized distribution rather than on a hard-truncated and renormalized distribution. 
This distinction is important: hard truncation discards the probability mass outside $S$, whereas tail compensation preserves this mass as an additional bucket.

\paragraph{Coarse-graining view.}
Let
\begin{equation}
m=\frac{1}{2}(p+q),
\quad
\tilde{m}=\frac{1}{2}(\tilde{p}+\tilde{q}).
\end{equation}
The tail mass of $m$ is
\begin{equation}
m_{\mathrm{tail}}
=
\sum_{v\in\bar{S}}m(v)
=
\frac{1}{2}
\left(
p_{\mathrm{tail}}+q_{\mathrm{tail}}
\right).
\end{equation}
The full-vocabulary JSD is
\begin{equation}
D_{\mathrm{JSD}}(q,p)
=
\frac{1}{2}D_{\mathrm{KL}}(q\|m)
+
\frac{1}{2}D_{\mathrm{KL}}(p\|m),
\end{equation}
while the Top-K JSD with tail compensation is
\begin{equation}
D_{\mathrm{TopK\mbox{-}JSD}}(q,p)
=
D_{\mathrm{JSD}}(\tilde{q},\tilde{p})
=
\frac{1}{2}D_{\mathrm{KL}}(\tilde{q}\|\tilde{m})
+
\frac{1}{2}D_{\mathrm{KL}}(\tilde{p}\|\tilde{m}).
\end{equation}

We next decompose the difference between full JSD and Top-K JSD. 
Assume $q_{\mathrm{tail}}>0$, $p_{\mathrm{tail}}>0$, and $m_{\mathrm{tail}}>0$. 
Define the conditional tail distributions
\begin{equation}
q_{\bar{S}}(v)
=
\frac{q(v)}{q_{\mathrm{tail}}},
\quad
p_{\bar{S}}(v)
=
\frac{p(v)}{p_{\mathrm{tail}}},
\quad
m_{\bar{S}}(v)
=
\frac{m(v)}{m_{\mathrm{tail}}},
\quad
v\in\bar{S}.
\end{equation}
By the chain rule of KL divergence under a partition $\{S,\bar{S}\}$, we have
\begin{equation}
D_{\mathrm{KL}}(q\|m)
=
D_{\mathrm{KL}}(\tilde{q}\|\tilde{m})
+
q_{\mathrm{tail}}
D_{\mathrm{KL}}(q_{\bar{S}}\|m_{\bar{S}}),
\label{eq:app_kl_decompose_q}
\end{equation}
and
\begin{equation}
D_{\mathrm{KL}}(p\|m)
=
D_{\mathrm{KL}}(\tilde{p}\|\tilde{m})
+
p_{\mathrm{tail}}
D_{\mathrm{KL}}(p_{\bar{S}}\|m_{\bar{S}}).
\label{eq:app_kl_decompose_p}
\end{equation}
Substituting Eq.~\eqref{eq:app_kl_decompose_q} and Eq.~\eqref{eq:app_kl_decompose_p} into the definition of JSD gives
\begin{align}
D_{\mathrm{JSD}}(q,p)
&=
D_{\mathrm{JSD}}(\tilde{q},\tilde{p})
\nonumber\\
&\quad+
\frac{1}{2}
q_{\mathrm{tail}}
D_{\mathrm{KL}}(q_{\bar{S}}\|m_{\bar{S}})
+
\frac{1}{2}
p_{\mathrm{tail}}
D_{\mathrm{KL}}(p_{\bar{S}}\|m_{\bar{S}}).
\label{eq:app_jsd_decompose}
\end{align}
Therefore,
\begin{equation}
D_{\mathrm{TopK\mbox{-}JSD}}(q,p)
=
D_{\mathrm{JSD}}(\tilde{q},\tilde{p})
\le
D_{\mathrm{JSD}}(q,p).
\label{eq:app_topk_jsd_lower_bound}
\end{equation}
The approximation error is
\begin{equation}
\Delta_{\mathrm{tail}}
=
D_{\mathrm{JSD}}(q,p)
-
D_{\mathrm{TopK\mbox{-}JSD}}(q,p)
=
\frac{1}{2}
q_{\mathrm{tail}}
D_{\mathrm{KL}}(q_{\bar{S}}\|m_{\bar{S}})
+
\frac{1}{2}
p_{\mathrm{tail}}
D_{\mathrm{KL}}(p_{\bar{S}}\|m_{\bar{S}}).
\label{eq:app_tail_error}
\end{equation}
Thus, Top-K JSD with tail compensation preserves the exact JSD contribution on the selected support $S$ and only discards the fine-grained divergence inside the aggregated tail region $\bar{S}$. 
The total probability mass of the tail is still preserved by $p_{\mathrm{tail}}$ and $q_{\mathrm{tail}}$.

\paragraph{Error interpretation.}
Eq.~\eqref{eq:app_tail_error} shows that the approximation error depends on two factors: the tail masses $p_{\mathrm{tail}}$ and $q_{\mathrm{tail}}$, and the internal discrepancy between the conditional tail distributions. 
When the selected Top-K support captures most of the probability mass, $p_{\mathrm{tail}}$ and $q_{\mathrm{tail}}$ are small, and the residual term is correspondingly limited. 
Moreover, the high-probability tokens in $S$ are kept explicitly, so the dominant token-level alignment signal is retained.

This property differs from hard truncation. 
If the tail is simply removed and the remaining probabilities are renormalized, the resulting objective changes the relative scale of the retained tokens and ignores the mismatch in the total omitted probability mass. 
In contrast, tail compensation preserves the mass outside $S$:
\begin{equation}
\tilde{p}_{\mathrm{tail}}=p_{\mathrm{tail}},
\quad
\tilde{q}_{\mathrm{tail}}=q_{\mathrm{tail}},
\end{equation}
so the objective still penalizes cases where the student and teacher assign substantially different total probability to the omitted vocabulary.

\paragraph{Memory complexity.}
Full-vocabulary token-level distillation requires storing or computing distributions over all vocabulary tokens for each valid response position. 
For batch size $B$, response length $T$, and vocabulary size $V$, the memory complexity is
\begin{equation}
O(BTV).
\end{equation}
Top-K JSD keeps only the union of two Top-K supports and one tail bucket. 
The resulting complexity becomes
\begin{equation}
O(BTK),
\quad
K\ll V,
\end{equation}
up to a small constant factor from the union support and the tail bucket. 
Therefore, Top-K JSD with tail compensation is a mass-preserving coarse-grained approximation to full-vocabulary JSD, retaining the dominant distributional alignment signal while substantially reducing the memory overhead of online token-level distillation.

\subsection{Optimization Rationale for Applying PTD to Failed Trajectories}
\label{app:Optimizatioin_rationale}

We analyze why PTD is applied to failed trajectories rather than all sampled trajectories. 
For a question $x$, let the policy sample a group of responses $\{y_i\}_{i=1}^{G}$ with binary verifiable rewards $r_i\in\{0,1\}$. 
The group-relative advantage used by GRPO is
\begin{equation}
A_i
=
\frac{
r_i-\bar{r}
}{
\sigma_r+\epsilon
},
\quad
\bar{r}
=
\frac{1}{G}
\sum_{j=1}^{G}r_j,
\quad
\sigma_r
=
\operatorname{std}
\left(
\{r_j\}_{j=1}^{G}
\right).
\label{eq:app_group_advantage}
\end{equation}

\paragraph{Degeneration under all-fail groups.}
When all sampled trajectories fail, we have
\begin{equation}
r_1=r_2=\cdots=r_G=0.
\end{equation}
Then
\begin{equation}
\bar{r}=0,
\quad
\sigma_r=0,
\quad
A_i=0,\quad \forall i.
\label{eq:app_all_fail_advantage}
\end{equation}
Thus, the clipped policy-gradient term in GRPO provides no discriminative improvement signal for the failed trajectories. 
Ignoring the KL regularization term, the token-level GRPO gradient for a sampled token $y_{i,t}$ is proportional to
\begin{equation}
-A_i
\nabla_{\theta}
\log
\pi_{\theta}
\left(
y_{i,t}
\mid
x,y_{i,<t}
\right),
\end{equation}
which vanishes when $A_i=0$. 
Therefore, all-fail groups require an additional corrective signal that can provide token-level guidance beyond scalar outcome rewards.

\paragraph{Negative advantage is not equivalent to corrective supervision.}
When a group contains both successful and failed responses, let
\begin{equation}
m=\sum_{i=1}^{G}r_i,
\quad
0<m<G.
\end{equation}
For a failed trajectory with $r_i=0$, its advantage is
\begin{equation}
A_i
=
\frac{
-\bar{r}
}{
\sigma_r+\epsilon
}
<0.
\label{eq:app_failed_negative_advantage}
\end{equation}
For a successful trajectory with $r_i=1$, its advantage is
\begin{equation}
A_i
=
\frac{
1-\bar{r}
}{
\sigma_r+\epsilon
}
>0.
\label{eq:app_success_positive_advantage}
\end{equation}
The negative advantage in Eq.~\eqref{eq:app_failed_negative_advantage} discourages the sampled failed trajectory. 
However, it does not specify which alternative tokens should receive higher probability at each intermediate reasoning step. 
In other words, GRPO can suppress low-reward sampled actions, but it does not directly construct a token-level corrective distribution for the failed reasoning path.

PTD supplies such a distributional correction. 
For a failed trajectory, the PTD gradient is induced by
\begin{equation}
D_{\mathrm{PTD}}
\left(
q^h_{i,t},
p_{i,t}
\right),
\quad
q^h_{i,t}
=
\pi_{\mathrm{ref}}
\left(
\cdot
\mid
x^h,y_{i,<t}
\right),
\quad
p_{i,t}
=
\pi_{\theta}
\left(
\cdot
\mid
x,y_{i,<t}
\right).
\end{equation}
Thus, PTD complements the scalar reward signal with a dense token-level target generated from the hint-conditioned self-teacher.

\paragraph{Over-regularization on successful trajectories.}
Applying PTD to successful trajectories is not always desirable. 
A successful trajectory already receives a positive reward-driven update from GRPO, as shown in Eq.~\eqref{eq:app_success_positive_advantage}. 
If PTD is additionally applied to this trajectory, the optimization objective contains
\begin{equation}
\lambda_{\mathrm{PTD}}
\sum_{t=1}^{T_i}
m_{i,t}
D_{\mathrm{PTD}}
\left(
q^h_{i,t},
p_{i,t}
\right),
\quad
r_i=1.
\label{eq:app_success_ptd_term}
\end{equation}
Since $q^h_{i,t}$ is conditioned on the privileged context $x^h=[x;h]$, it is not guaranteed to coincide with the question-only distribution that produced a correct trajectory. 
Therefore, Eq.~\eqref{eq:app_success_ptd_term} may pull a reward-consistent on-policy behavior toward a hint-conditioned distribution. 
This introduces an additional regularization bias that is not required for solving the instance.

This can be made explicit by decomposing an all-trajectory PTD objective into failed and successful contributions with a shared normalization. Let $N_{\mathrm{fail}}$ and $N_{\mathrm{succ}}$ denote the numbers of valid response tokens in failed and successful trajectories, respectively, and let $N=N_{\mathrm{fail}}+N_{\mathrm{succ}}$. Then:
\begin{equation}
\mathcal{L}_{\mathrm{PTD}}^{\mathrm{all}}
=
\mathcal{L}_{\mathrm{PTD}}^{\mathrm{fail}}
+
\mathcal{L}_{\mathrm{PTD}}^{\mathrm{succ}},
\end{equation}
where
\begin{equation}
\mathcal{L}_{\mathrm{PTD}}^{\mathrm{fail}}
=
\frac{
\sum_{i:r_i=0}
\sum_{t=1}^{T_i}
m_{i,t}
D_{\mathrm{PTD}}
\left(
q^h_{i,t},
p_{i,t}
\right)
}{
N+\epsilon
},
\end{equation}
and
\begin{equation}
\mathcal{L}_{\mathrm{PTD}}^{\mathrm{succ}}
=
\frac{
\sum_{i:r_i=1}
\sum_{t=1}^{T_i}
m_{i,t}
D_{\mathrm{PTD}}
\left(
q^h_{i,t},
p_{i,t}
\right)
}{
N+\epsilon
}.
\end{equation}
The failed part provides corrective supervision for trajectories with insufficient reward feedback. 
The successful part instead constrains already correct trajectories to match the hint-conditioned self-teacher. 
Since the teacher and student are conditioned on different contexts, this term can conflict with the on-policy reward-improvement direction.

\paragraph{Selective PTD objective.}
PTD-PO therefore applies distillation only to failed trajectories in PTD-activated groups. 
Let
\begin{equation}
\alpha_i
=
\mathbf{1}
\left[
\bar{r}(x)<\tau_{\mathrm{ptd}}
\right]
\mathbf{1}
\left[
r_i=0
\right],
\label{eq:app_ptd_indicator}
\end{equation}
where $\tau_{\mathrm{ptd}}$ is the group-level activation threshold. 
The selective PTD objective can be written as
\begin{equation}
\mathcal{L}_{\mathrm{PTD}}
=
\frac{
\sum_{i=1}^{G}
\alpha_i
\sum_{t=1}^{T_i}
m_{i,t}
D_{\mathrm{PTD}}
\left(
q^h_{i,t},
p_{i,t}
\right)
}{
\sum_{i=1}^{G}
\alpha_i
\sum_{t=1}^{T_i}
m_{i,t}
+\epsilon
}.
\label{eq:app_selective_ptd}
\end{equation}
With $\tau_{\mathrm{ptd}}=1.0$, PTD is activated for groups that contain at least one failed trajectory, while fully correct groups are excluded from distillation. 
Within activated groups, only failed trajectories are distilled.

The final objective is
\begin{equation}
\mathcal{L}_{\mathrm{PTD\mbox{-}GRPO}}
=
-J_{\mathrm{GRPO}}(\theta)
+
\lambda_{\mathrm{PTD}}
\mathcal{L}_{\mathrm{PTD}}.
\label{eq:app_selective_ptd_final}
\end{equation}
This objective preserves the reward-driven update on successful trajectories and injects privileged token-level correction only where the outcome reward is insufficient.

\paragraph{Gradient decomposition.}
Let
\begin{equation}
g_{\mathrm{GRPO}}
=
\nabla_{\theta}
\left(
-J_{\mathrm{GRPO}}
\right),
\quad
g_{\mathrm{PTD}}
=
\nabla_{\theta}
\mathcal{L}_{\mathrm{PTD}}.
\end{equation}
The total gradient is
\begin{equation}
g_{\mathrm{total}}
=
g_{\mathrm{GRPO}}
+
\lambda_{\mathrm{PTD}}
g_{\mathrm{PTD}}.
\end{equation}
Under all-trajectory distillation, the PTD gradient can be decomposed as
\begin{equation}
g_{\mathrm{PTD}}^{\mathrm{all}}
=
g_{\mathrm{PTD}}^{\mathrm{fail}}
+
g_{\mathrm{PTD}}^{\mathrm{succ}}.
\end{equation}
The term $g_{\mathrm{PTD}}^{\mathrm{fail}}$ provides a corrective gradient for failed trajectories. 
The term $g_{\mathrm{PTD}}^{\mathrm{succ}}$ imposes hint-conditioned regularization on successful trajectories and is not guaranteed to align with $g_{\mathrm{GRPO}}$. 
With the default $\tau_{\mathrm{ptd}}=1.0$, selective PTD removes the successful-trajectory contribution and renormalizes over failed tokens:
\begin{equation}
g_{\mathrm{PTD}}^{\mathrm{selective}}
=
\frac{N+\epsilon}{N_{\mathrm{fail}}+\epsilon}
g_{\mathrm{PTD}}^{\mathrm{fail}}.
\end{equation}
Therefore, applying PTD to failed trajectories reduces unnecessary regularization on reward-consistent behaviors while preserving dense correction for trajectories where GRPO provides weak or incomplete token-level supervision.

\section{Additional Experiments and Analysis}
\label{app:more_exp}

\subsection{Compatibility With Other RLVR Optimizers}
\label{app:compatiable_rlvr}

\makeatletter
\AddToHookNext{cmd/@vtryfc/before}{%
  \setlength{\@fptop}{0pt}%
  \setlength{\@fpsep}{18pt}%
  \setlength{\@fpbot}{0pt plus 1fil}%
}
\makeatother

\begin{table*}[htp]
\centering
\caption{Optimizer compatibility and comparison with open-source multimodal models.}
\label{tab:ablation_rl_ptd}

\Large
\renewcommand{\arraystretch}{1.2}
\setlength{\tabcolsep}{1.3mm}

\resizebox{\textwidth}{!}{%
\begin{tabular}{l|cccccc|ccccc|cc}
\toprule
\multirow{2}{*}{\textbf{Method}} 
& \multicolumn{6}{c|}{\textbf{General Multimodal Reasoning}} 
& \multicolumn{5}{c|}{\textbf{Vision-Dependent Multimodal Reasoning}} 
& \multicolumn{2}{c}{\textbf{Overall}} \\
\cmidrule(lr){2-7} \cmidrule(lr){8-12} \cmidrule(lr){13-14}
& \textbf{MMK12} 
& \textbf{Geo3K} 
& \textbf{MathVerse} 
& \textbf{MathVista} 
& \textbf{We-Math} 
& \textbf{AVG} 
& \textbf{MMMU-Pro} 
& \textbf{Counting} 
& \textbf{MathVerse$_V$} 
& \textbf{LogicVista} 
& \textbf{AVG} 
& \textbf{AVG} 
& \textbf{$\Delta_{rel}^{\%}$} \\
\midrule

\multicolumn{14}{c}{\textbf{Open-Source Methods and Models}} \\
\midrule

Qwen2.5-VL-7B
& 49.40 & 42.90 & 40.20 & 68.50 & 47.80 & 49.76
& 36.40 & 76.50 & 36.60 & 45.20 & 48.68
& 49.28 & -- \\

ThinkLite-VL-7B
& 50.50 & 45.30 & 33.70 & 73.60 & 43.70 & 49.36
& 32.30 & 88.50 & 30.90 & 44.30 & 49.00
& 49.20 & -- \\

VL-Rethinker-7B
& 66.00 & 44.40 & 68.80 & 65.20 & 67.50 & 62.38
& 37.60 & 76.50 & 64.80 & 47.00 & 56.48
& 59.76 & -- \\

MM-Eureka-7B
& 61.70 & 41.10 & 66.70 & 66.30 & 66.60 & 60.48
& 40.60 & 79.00 & 63.60 & 47.90 & 57.78
& 59.28 & -- \\

NoisyRollout-7B$^{*}$
& 51.20 & 55.20 & 58.80 & 72.90 & 64.40 & 60.50
& 38.20 & 88.50 & 51.70 & 47.00 & 56.35
& 58.66 & -- \\

PAPO$_D$-7B
& 80.80 & 48.80 & 69.50 & 72.30 & 69.40 & 68.16
& 40.20 & 89.00 & 66.60 & 45.90 & 60.43
& 64.72 & -- \\

VPPO-RL-7B
& 81.80 & 46.90 & 71.10 & 70.50 & 70.60 & 68.18
& 40.20 & 88.50 & 67.60 & 48.80 & 61.28
& 65.11 & -- \\

VGPO-7B
& 81.50 & 45.80 & 71.60 & 74.10 & 72.50 & 69.10
& 40.50 & 95.50 & 67.60 & 49.40 & 63.25
& 66.50 & -- \\

\midrule

\multicolumn{14}{c}{\textbf{Qwen3-VL-2B-Thinking}} \\
\midrule

Qwen3-VL-2B
& 12.32 & 25.23 & 26.74 & 44.35 & 27.08& 27.14& 13.61& 80.88 & 24.96 & 18.57 & 34.51 & 30.42& -- \\


GRPO
& 47.71 & 39.29 & 47.98 & 53.58 & 57.12 & 49.13
& 20.51 & 80.13 & 45.41 & 29.84 & 43.97
& 46.84 & -- \\

PAPO$_G$
& 48.57 & 41.08 & 51.89 & 56.08 & 59.17 & 51.36
& 23.42 & 80.63 & 50.05 & 32.83 & 46.73
& 49.30 & $\uparrow$5.25 \\

CFPO$_G$
& 48.04 & 44.80 & 61.92 & \textbf{66.96} & 68.25 & 57.99
& 27.96 & 82.50 & 58.43 & 44.45 & 53.34
& 55.92 & $\uparrow$19.39 \\

\textbf{PTD-PO$_{GRPO}$}
& \textbf{53.41} & \textbf{46.38} & \textbf{66.82} & 60.44 & \textbf{70.51} & \textbf{59.51}
& \textbf{29.97} & \textbf{87.75} & \textbf{63.09} & \textbf{49.61} & \textbf{57.61}
& \textbf{58.66} & \textbf{$\uparrow$25.23} \\

\midrule

\multicolumn{14}{c}{\textbf{Qwen3-VL-4B-Thinking}} \\
\midrule

Qwen3-VL-4B
& 37.31 & 48.88 & 56.28 & 61.54 & 52.95 & 51.39 & 30.29 & 91.38 & 53.01 & 39.32 & 53.5 & 52.33 & -- \\

GRPO 
& 65.92 & 58.15 & 74.38 & 67.33 & 75.69 & 68.29
& 35.74 & 92.25 & 70.30 & 52.01 & 62.58
& 65.75 & -- \\

\textbf{PTD-PO$_{GRPO}$}
& 71.33 & 61.02 & 77.56 & 69.55 & 79.39 & 71.77
& 38.56 & 92.88 & 74.16 & 57.63 & 65.81
& 69.12 & \textbf{$\uparrow$5.13} \\

DAPO 
& 64.03 & 53.72 & 70.34 & 67.39 & 73.46 & 65.79
& 35.01 & 92.25 & 66.74 & 52.29 & 61.57
& 63.91 & -- \\

\textbf{PTD-PO$_{DAPO}$}
& 67.51 & 56.03 & 73.85 & 69.01 & 74.97 & 68.27
& 35.14 & 92.62 & 70.50 & 52.82 & 62.77
& 65.83 & \textbf{$\uparrow$3.00} \\

GSPO 
& 70.43 & 60.77 & 78.51 & 69.58 & 79.46 & 71.75
& 38.75 & 92.62 & 74.25 & 59.06 & 66.17
& 69.27 & -- \\

\textbf{PTD-PO$_{GSPO}$} 
& \textbf{71.82} & \textbf{61.63} & \textbf{79.12} & \textbf{73.89} & \textbf{79.86} & \textbf{73.26}
& \textbf{40.30} & \textbf{92.94} & \textbf{75.23} & \textbf{60.65} & \textbf{67.28}
& \textbf{70.60} & \textbf{$\uparrow$1.92} \\

\bottomrule
\end{tabular}
}
\end{table*}

In the Qwen3-VL-4B-Thinking group of Table~\ref{tab:ablation_rl_ptd}, PTD-PO$_{GRPO}$, PTD-PO$_{DAPO}$, and PTD-PO$_{GSPO}$ denote the PTD-augmented variants of GRPO \citep{shao2024deepseekmath}, DAPO \citep{yu2026dapo}, and GSPO \citep{zheng2025group}, respectively; each retains its underlying optimizer and adds PTD as an auxiliary distillation objective.
Adding PTD yields relative overall gains of $5.13\%$, $3.00\%$, and $1.92\%$ over the respective optimizer baselines, showing that PTD can provide dense corrective guidance to failed trajectories as an auxiliary distillation module across RLVR objectives.

Table~\ref{tab:ablation_rl_ptd} also compares PTD with open-source models and methods across nine multimodal reasoning benchmarks: Qwen2.5-VL \citep{bai2025qwen25}, Qwen3-VL \citep{bai2025qwen3}, ThinkLite-VL \citep{wang2025thinklite}, VL-Rethinker \citep{wang2025vlrethinker}, MM-Eureka \citep{meng2025mm}, NoisyRollout \citep{liu2025noisyrollout}, PAPO \citep{wang2025perception}, VPPO-RL \citep{huang2025spotlight}, VGPO \citep{wang2026vgpo}, and CFPO \citep{yu2026cfpo}.
AVG denotes the mean over the five general, four vision-dependent, or all nine benchmarks; comparison averages are recomputed over these columns.
Bold highlights the best scores within each backbone group, and $\Delta_{rel}^{\%}$ denotes relative overall improvement over GRPO within the same group.
The asterisk marks NoisyRollout's training on Geo3K.
Differences in backbones, training data, and evaluation protocols mean these cross-model results provide broader context rather than isolating the effect of optimization.

\subsection{Sensitivity Analysis on the Top-$K$ Support Size}
\label{app:abl_Top_K}

\begin{table}[htp]
\centering
\caption{Effect of the Top-$K$ support size in PTD on Qwen3-VL-4B- and 8B-Thinking.}
\label{tab:ablation_topk}

\large
\renewcommand{\arraystretch}{1.15}
\setlength{\tabcolsep}{3.5mm}

\resizebox{0.85\textwidth}{!}{%
\begin{tabular}{l|c|cccc}
\toprule
\textbf{Model} 
& \textbf{Top-$K$} 
& \textbf{MMK12} 
& \textbf{MathVerse} 
& \textbf{MMMU-Pro} 
& \textbf{LogicVista} \\
\midrule

\multirow{4}{*}{Qwen3-VL-4B-Thinking}
& 50  & 69.73 & 77.32 & 37.98 & 57.24 \\
& 100 & 71.33 & 77.56 & 38.56 & 57.63 \\
& 200 & 71.90 & 78.11 & 37.67 & 57.70 \\

\midrule

\multirow{4}{*}{Qwen3-VL-8B-Thinking}
& 50  & 73.84 & 77.20 & 41.77 & 58.94 \\
& 100 & 76.55 & 80.14 & 44.62 & 61.13 \\
& 200 & 76.59 & 80.67 & 44.10 & 61.38 \\

\bottomrule
\end{tabular}
}
\end{table}

\begin{figure*}[tp]
    \vspace{5pt}
    \centering
    \includegraphics[width=\textwidth]{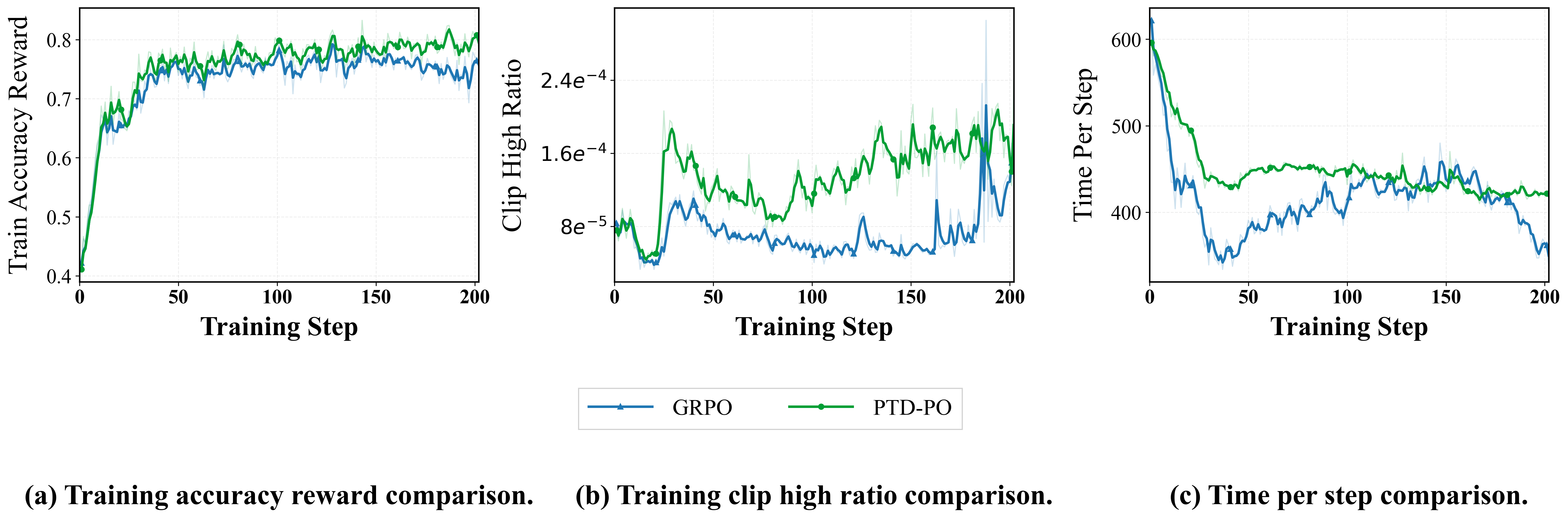}
    \caption{Training dynamics and efficiency comparisons.}
    \label{fig:training_accuracy_time}
\end{figure*}

Table~\ref{tab:ablation_topk} studies the effect of the Top-$K$ support size used in the PTD divergence. 
We vary $K$ while keeping all other training settings unchanged, and report results on four representative benchmarks covering both general reasoning and vision-dependent reasoning. 
This experiment examines whether the default choice $K=100$ provides a good trade-off between retaining informative token-level distribution signals and reducing the memory cost of full-vocabulary distillation.


\subsection{Training Dynamics and Efficiency Analysis}
\label{app:training_curves}
Figure~\ref{fig:training_accuracy_time} compares the training dynamics of GRPO and PTD-PO on the Qwen3-VL-4B-Thinking model. 
After the early warm-up stage, PTD-PO not only maintains a consistently higher training accuracy reward, but also continues to improve in the later training phase, whereas standard GRPO gradually enters a plateau. 
This suggests that the privileged tutoring signal keeps providing corrective guidance for failed rollouts and enables more sustained learning beyond the sparse reward signal. 
The clip high ratio of PTD-PO is higher than that of GRPO throughout most of training, indicating that the policy receives stronger update signals and preserves a higher degree of exploration under privileged guidance. 
In terms of efficiency, PTD-PO has a comparable time-per-step trend to GRPO after the initial stage, with only moderate overhead from teacher-side distribution computation. 
These results show that PTD-PO improves training effectiveness on the 4B model while maintaining stable optimization behavior and practical training efficiency.

\subsection{Detailed Ablation Results}
\label{app:detailed_ablation}

Table~\ref{tab:ablation_PID-GRPO} provides the complete numerical results underlying Figure~\ref{fig:ptd_ablation}, including per-benchmark scores, category averages, and relative improvements.
AVG denotes the mean over the five general, four vision-dependent, or all nine benchmarks; $\Delta_{rel}^{\%}$ is relative to GRPO within the same backbone.
The label \texttt{thr} denotes $\tau_{\mathrm{ptd}}$; the threshold variants use WSPH and distill only failed trajectories, while \textit{All Trajectories} also distills correct rollouts.
The regular-hint variant uses $\tau_{\mathrm{ptd}}=1.0$ but removes the visual-grounding and trap-filtering constraints. Bold marks the default setting, $\tau_{\mathrm{ptd}}=1.0$ with WSPH.

\begin{table*}[htbp]
\centering
\caption{
Detailed ablation results for PTD-PO.
}
\label{tab:ablation_PID-GRPO}

\Large
\renewcommand{\arraystretch}{1.2}
\setlength{\tabcolsep}{1.2mm}

\resizebox{\textwidth}{!}{%
\begin{tabular}{l|ccccc|cc|cccc|cc|cc}
\toprule
\multirow{2}{*}{\textbf{Method / Setting}} 
& \multicolumn{7}{c|}{\textbf{General Multimodal Reasoning}} 
& \multicolumn{6}{c|}{\textbf{Vision-Dependent Multimodal Reasoning}} 
& \multicolumn{2}{c}{\textbf{Overall}} \\
\cmidrule(lr){2-8} \cmidrule(lr){9-14} \cmidrule(lr){15-16}
& \textbf{MMK12} 
& \textbf{Geo3K} 
& \textbf{MathVerse} 
& \textbf{MathVista} 
& \textbf{We-Math} 
& \textbf{AVG} 
& \textbf{$\Delta_{rel}^{\%}$} 
& \textbf{MMMU-Pro} 
& \textbf{Counting} 
& \textbf{MathVerse$_V$} 
& \textbf{LogicVista} 
& \textbf{AVG} 
& \textbf{$\Delta_{rel}^{\%}$} 
& \textbf{AVG} 
& \textbf{$\Delta_{rel}^{\%}$} \\
\midrule

\multicolumn{16}{c}{\textbf{Qwen3-VL-2B-Thinking}} \\
\midrule
GRPO 
& 47.71 & 46.34 & 61.06 & 58.41 & 57.33 & 54.17 & -- 
& 19.63 & 82.56 & 57.27 & 41.50 & 50.24 & -- 
& 52.42 & -- \\

+PTD(thr 0.2) 
& 45.34 & 48.63 & 65.38 & 60.42 & 69.81 & 57.92 & $\uparrow$6.92
& 27.17 & 86.69 & 61.74 & 45.47 & 55.27 & $\uparrow$10.01
& 56.74 & $\uparrow$8.24 \\

+PTD(thr 0.4) 
& 43.59 & 46.65 & 63.62 & 60.74 & 68.23 & 56.57 & $\uparrow$4.42
& 27.57 & 90.31 & 60.21 & 46.92 & 56.25 & $\uparrow$11.97
& 56.43 & $\uparrow$7.64 \\

\textbf{+PTD(thr 1.0)} 
& \textbf{52.22} & \textbf{45.90} & \textbf{66.81} & \textbf{60.84} & \textbf{70.19} & \textbf{59.19} & \textbf{$\uparrow$9.27}
& \textbf{29.60} & \textbf{85.69} & \textbf{62.67} & \textbf{48.52} & \textbf{56.62} & \textbf{$\uparrow$12.70}
& \textbf{58.05} & \textbf{$\uparrow$10.73} \\

+PTD(All Trajectories) 
& 53.14 & 45.78 & 66.50 & 60.46 & 69.69 & 59.11 & $\uparrow$9.13
& 29.59 & 84.12 & 62.78 & 49.50 & 56.50 & $\uparrow$12.46
& 57.95 & $\uparrow$10.54 \\

\midrule

\multicolumn{16}{c}{\textbf{Qwen3-VL-4B-Thinking}} \\
\midrule
GRPO 
& 65.92 & 58.15 & 74.38 & 67.33 & 75.69 & 68.29 & -- 
& 35.74 & 92.25 & 70.30 & 52.01 & 62.58 & -- 
& 65.75 & -- \\

+PTD(thr 0.2) 
& 66.88 & 60.77 & 75.89 & 67.81 & 76.05 & 69.48 & $\uparrow$1.74
& 37.32 & 92.88 & 71.89 & 54.64 & 64.18 & $\uparrow$2.57
& 67.13 & $\uparrow$2.09 \\

+PTD(thr 0.4) 
& 68.04 & 61.88 & 75.99 & 68.97 & 77.20 & 70.42 & $\uparrow$3.11
& 37.82 & 93.81 & 72.09 & 55.40 & 64.78 & $\uparrow$3.52
& 67.91 & $\uparrow$3.28 \\

\textbf{+PTD(thr 1.0)} 
& \textbf{71.33} & \textbf{61.02} & \textbf{77.56} & \textbf{69.55} & \textbf{79.39} & \textbf{71.77} & \textbf{$\uparrow$5.10}
& \textbf{38.56} & \textbf{92.88} & \textbf{74.16} & \textbf{57.63} & \textbf{65.81} & \textbf{$\uparrow$5.17}
& \textbf{69.12} & \textbf{$\uparrow$5.13} \\

+PTD(All Trajectories) 
& 68.68 & 59.05 & 75.40 & 69.45 & 77.95 & 70.11 & $\uparrow$2.65
& 37.19 & 92.75 & 71.43 & 55.23 & 64.15 & $\uparrow$2.52
& 67.46 & $\uparrow$2.60 \\

+PTD(regular hints) 
& 69.03 & 60.03 & 77.13 & 68.39 & 79.22 & 70.76 & $\uparrow$3.62
& 37.38 & 91.38 & 72.89 & 54.95 & 64.15 & $\uparrow$2.52
& 67.82 & $\uparrow$3.15 \\

\midrule

\multicolumn{16}{c}{\textbf{Qwen3-VL-8B-Thinking}} \\
\midrule
GRPO 
& 73.18 & 61.02 & 78.12 & 71.67 & 80.37 & 72.87 & -- 
& 43.06 & 90.81 & 74.69 & 58.47 & 66.76 & -- 
& 70.15 & -- \\

+PTD(thr 0.2) 
& 71.82 & 61.34 & 76.55 & 72.65 & 78.13 & 72.10 & $\downarrow$1.06
& 42.01 & 91.69 & 71.42 & 55.43 & 65.14 & $\downarrow$2.43
& 69.00 & $\downarrow$1.64 \\

+PTD(thr 0.4) 
& 72.69 & 61.71 & 77.19 & 73.62 & 79.86 & 73.01 & $\uparrow$0.19
& 42.36 & 92.25 & 73.54 & 56.04 & 66.05 & $\downarrow$1.06
& 69.92 & $\downarrow$0.34 \\

\textbf{+PTD(thr 1.0)} 
& \textbf{76.55} & \textbf{65.52} & \textbf{80.14} & \textbf{72.79} & \textbf{82.92} & \textbf{75.58} & \textbf{$\uparrow$3.72}
& \textbf{44.62} & \textbf{92.75} & \textbf{77.32} & \textbf{61.13} & \textbf{68.96} & \textbf{$\uparrow$3.29}
& \textbf{72.64} & \textbf{$\uparrow$3.55} \\

+PTD(All Trajectories) 
& 74.34 & 64.39 & 78.62 & 72.10 & 81.14 & 74.12 & $\uparrow$1.71
& 42.83 & 92.12 & 75.36 & 58.50 & 67.20 & $\uparrow$0.67
& 71.04 & $\uparrow$1.27 \\

+PTD(regular hints) 
& 72.51 & 60.73 & 77.39 & 71.00 & 80.16 & 72.36 & $\downarrow$0.70
& 41.62 & 92.25 & 73.66 & 56.35 & 65.97 & $\downarrow$1.18
& 69.52 & $\downarrow$0.90 \\

\bottomrule
\end{tabular}
}
\end{table*}

\clearpage
\section{Case Study}
\label{app:case_study}

Figure~\ref{fig:training_hint_cases} shows three selected training samples with original images and WSPH annotations; target answers are omitted.
These are training-data examples, not model-generated rollouts: only the frozen reference teacher sees WSPH, while the student receives the original question and image.

\definecolor{PTDHintBorder}{HTML}{268B79}
\definecolor{PTDHintBackground}{HTML}{EFF7F4}
\newcommand{\privilegedhintbox}[1]{%
\begingroup
\setlength{\fboxsep}{4pt}%
\setlength{\fboxrule}{0.6pt}%
\noindent\fcolorbox{PTDHintBorder}{PTDHintBackground}{%
\begin{minipage}{\dimexpr\linewidth-2\fboxsep-2\fboxrule\relax}
\raggedright
\textbf{\textcolor{PTDHintBorder}{WSPH.}}\par
#1
\end{minipage}}%
\endgroup}
\begin{figure}[!htbp]
\centering
\resizebox*{!}{0.77\textheight}{%
\begin{minipage}{0.98\textwidth}
\fontsize{9}{10}\selectfont
\textbf{(a) Geometry: correspondence under segment extensions}
\par\smallskip
\begin{minipage}[c]{0.20\linewidth}
\centering
\includegraphics[width=0.9\linewidth]{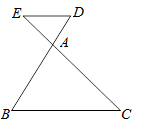}
\end{minipage}\hfill
\begin{minipage}[c]{0.76\linewidth}
\textbf{Question.} As shown in the figure, $DE \parallel BC$ intersects the extension of $BA$ at $D$ and the extension of $CA$ at $E$. Given that $AD = 4$ and $DE:BC = 1:2$, then $AB =$ \underline{\hspace{1em}}.
\end{minipage}
\par\smallskip
\privilegedhintbox{%
\begin{itemize}[leftmargin=1.2em,nosep,topsep=2pt]
\item Identify similar triangles formed by $DE \parallel BC$; focus on vertex correspondence at point A.
\item Ignore the direction of segment extensions; use DE:BC ratio to set proportional side relationships.
\item Relate AD to AB via similarity ratio---note AD corresponds to the segment opposite the same angle as AB.
\end{itemize}
}

\par\smallskip\hrule\smallskip
\textbf{(b) Statistical charts: selecting evidence across two views}
\par\smallskip
\centering
\includegraphics[width=0.58\linewidth]{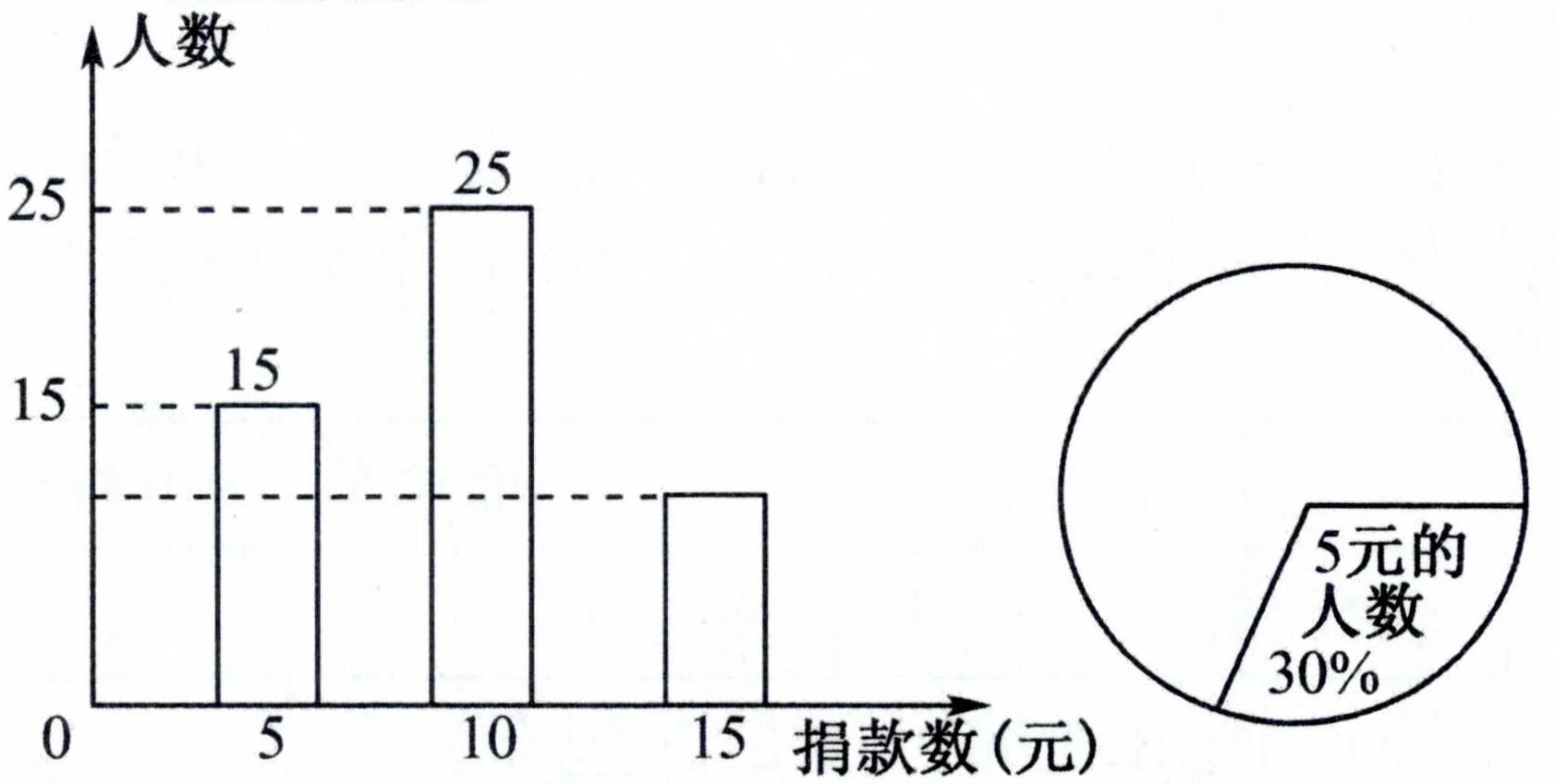}
\par
\raggedright
{\footnotesize Chart labels (translated): number of students; donation amount (yuan); students donating 5 yuan (30\%).\par}
\smallskip
\textbf{Question.} To welcome the arrival of `June 1st' Children's Day, students from a certain school participated in a love donation activity. A statistical chart of the donation amounts from a random sample of students is shown below. If the school has a total of 500 students, estimate the total donation amount of the students based on this sample.
\par\smallskip
\privilegedhintbox{%
\begin{itemize}[leftmargin=1.2em,nosep,topsep=2pt]
\item Use the 30\% segment and the 5 yuan bar count to determine the sample size.
\item Ignore the 15 yuan bar's height; calculate its count from the sample total minus other counts.
\item Compute the sample's average donation, then multiply by the school's total student count.
\end{itemize}
}

\par\smallskip\hrule\smallskip
\textbf{(c) Kinetic energy: comparing mass and speed across gas samples}
\par\smallskip
\begin{minipage}[c]{0.45\linewidth}
\centering
\includegraphics[width=\linewidth]{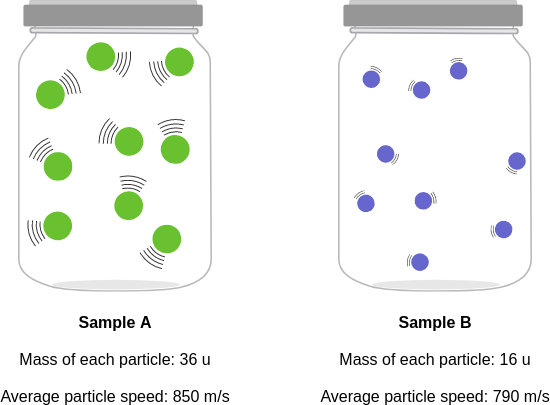}
\end{minipage}\hfill
\begin{minipage}[c]{0.51\linewidth}
\textbf{Question.} The diagrams below show two pure samples of gas in identical closed, rigid containers. Each colored ball represents one gas particle. Both samples have the same number of particles.
Compare the average kinetic energies of the particles in each sample. Which sample has the higher temperature?
\par\smallskip
A. neither; the samples have the same temperature\\
B. sample A\\
C. sample B
\end{minipage}
\par\smallskip
\privilegedhintbox{%
\begin{itemize}[leftmargin=1.2em,nosep,topsep=2pt]
\item Ignore the number of particles and container details; they are irrelevant for temperature comparison.
\item Calculate the product of each particle's mass and the square of its average speed for both samples.
\item Focus on comparing these products directly; the higher value indicates the higher temperature.
\end{itemize}
}
\end{minipage}
}
\caption{Original training examples with WSPH. The hints provide visual grounding and reasoning guidance, while final answers are omitted.}
\label{fig:training_hint_cases}
\end{figure}

\clearpage

\section{Limitations and Future Work}
\label{app:limitations}

\paragraph{Limitations.}
Although PTD-PO improves failed-trajectory learning through dense privileged guidance, it does not fully eliminate reasoning failures in highly compositional multimodal problems.
In addition, the relative gains of PTD-PO become less pronounced on larger models compared with smaller ones.
One possible reason is the mismatch between model capability and training-data difficulty: stronger models produce fewer failed trajectories during RLVR training, which reduces the number of cases where PTD can be activated to provide corrective supervision.
Since PTD-PO is mainly designed to improve learning from failed rollouts, its effective scope naturally depends on the availability of sufficiently challenging examples that expose such failures.
Therefore, PTD-PO should be viewed as an effective way to improve corrective supervision for failed rollouts, rather than a complete solution to all visual grounding and symbolic reasoning errors.
Constructing harder and more capability-matched multimodal RLVR training data may further enlarge the benefit of privileged-information distillation, especially for stronger models.

\paragraph{Future Work.}
An important future direction is to construct harder and more capability-matched RLVR training data for stronger LVLMs.
Since PTD-PO mainly provides corrective supervision for failed trajectories, larger models may benefit less when the current training data is not sufficiently challenging and produces too few informative failures.
Designing more difficult multimodal reasoning tasks, especially those requiring fine-grained visual grounding, symbolic verification, and long-chain reasoning, could expose more failure cases and further enlarge the effective scope of privileged-information distillation.
Beyond static multimodal reasoning, another promising direction is to extend PTD-style training to multimodal agent settings.
In agentic tasks, models need to interact with environments, use tools, update observations, and make long-horizon decisions.
Future work may construct non-goal-revealing privileged hints from intermediate observations, tool feedback, or expert action traces, so as to guide failed agent trajectories without directly revealing the final goal state.
Such an extension could help improve exploration, grounding, and recovery behaviors in multimodal agents.


\end{document}